\documentclass[final]{cvpr}
\usepackage[table]{xcolor}
\usepackage[numbers,sort,compress]{natbib}
\usepackage{times}
\usepackage{epsfig}
\usepackage{subcaption}
\usepackage{graphicx}
\usepackage{booktabs}
\usepackage{amsmath}
\usepackage{amssymb}
\newcommand{\E}{\mathbb{E}}

\newcommand{\minus}{\scalebox{0.5}[1.0]{$-$}}
\usepackage{adjustbox}
\usepackage{amsthm}

\DeclareMathOperator*{\argmax}{argmax}

\usepackage{stackengine}

\makeatletter
\newcommand{\distas}[1]{\mathbin{\overset{#1}{\kern\z@\sim}}}%

\newcommand{\beqs}{\vspace{0mm}\begin{eqnarray}}
\newcommand{\eeqs}{\vspace{0mm}\end{eqnarray}}
\newcommand{\barr}{\begin{array}}
\newcommand{\earr}{\end{array}}

\newcommand{\xv}{\boldsymbol{x}}

\newcommand{\ours}{{ASR-Norm}}

\newcommand{\betav}[0]{{\boldsymbol{\beta}}}
\newcommand{\gammav}[0]{{\boldsymbol{\gamma}} }

\newcommand{\muv}[0]{{\boldsymbol{\mu}}}

\newcommand{\sigmav}[0]{{\boldsymbol{\sigma}} }

\newtheorem{theorem}{Theorem}

\newtheorem{remark}[theorem]{Remark}

\usepackage[pagebackref=true,breaklinks=true,colorlinks,bookmarks=false]{hyperref}

\def\cvprPaperID{4346} 
\def\confYear{CVPR 2021}

\begin{document}

\title{Adversarially Adaptive Normalization for Single Domain
Generalization}
\author{Xinjie Fan$^{1,*}$, Qifei Wang$^{2}$, Junjie Ke$^{2}$, Feng Yang$^{2}$, Boqing Gong$^{2}$, Mingyuan Zhou$^1$\\
$^1$The University of Texas at Austin \quad $^2$Google Research
\\
\texttt{xfan@utexas.edu, \{qfwang, junjiek, fengyang, bgong\}@google.com}\\\texttt{mingyuan.zhou@mccombs.utexas.edu} }


\maketitle

\begin{abstract}
Single domain generalization aims to learn a model that performs well on many unseen domains with only one domain data for training. Existing works focus on studying the adversarial domain augmentation (ADA) to improve the model's generalization capability. The impact on domain generalization of the statistics of normalization layers is still underinvestigated. In this paper, we propose a generic normalization approach, {\it adaptive standardization and rescaling normalization} ({\ours}), to complement the missing part in previous works. {\ours} learns both the standardization and rescaling statistics via neural networks. 
This new form of normalization can be viewed as a generic form of the traditional normalizations. 
When trained with ADA, the statistics in ASR-Norm are learned to be 
adaptive to the data coming from different domains, and hence improves the model generalization performance across domains, especially on the target domain with large discrepancy from the source domain.
The experimental results show that {\ours} can bring consistent improvement to the state-of-the-art ADA approaches by 
$1.6\%$, $2.7\%$, and $6.3\%$ averagely
on the Digits, CIFAR-10-C, and PACS benchmarks, respectively. As a generic tool, the improvement introduced by {\ours} is agnostic to the choice of ADA methods. {\let\thefootnote\relax\footnote{{$^*$The main work was done during an internship at Google Research.}}} \vspace{-6mm}

\end{abstract}


\section{Introduction} 
\begin{figure}[tp!]
\centering
\includegraphics[height=4.9cm]{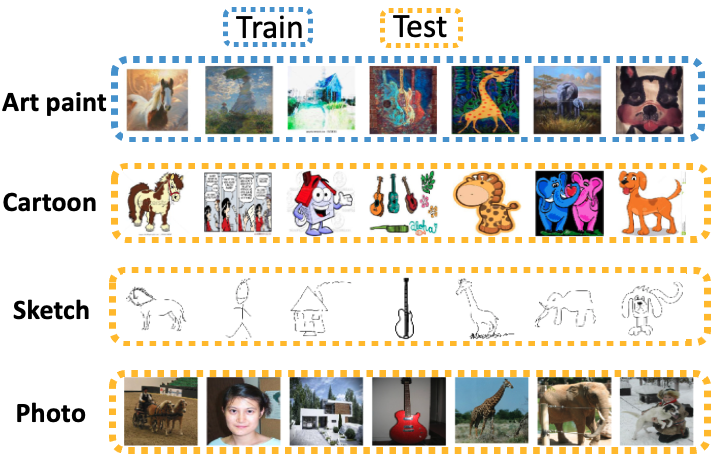}
\caption{Illustration of single domain generalization with the PACS \cite{li2017deeper} benchmark. The dataset contains $4$ domains: \emph{art paint}, \emph{cartoon}, \emph{sketch}, and \emph{photo} domains, which share the same categories that include \emph{dog}, \emph{elephant}, \emph{giraffe}, \emph{guitar}, \emph{house}, \emph{horse}, and \emph{person}. Single domain generalization aims at training a model on one source domain data (\emph{art paint} domain in the shown case), while generalizing well to other domains with very different visual presentations.} 
\label{fig:sdg_pacs} \vspace{-6.0mm}
\end{figure}

Deep learning has achieved remarkable success in various areas \cite{krizhevsky2017imagenet, lecun2015deep} where the training and test data are sampled from the same domain. In real applications, however, there is a great chance of applying a deep learning model to the data from a new domain unseen in the training dataset. The model that performs well on the training domain often cannot maintain a consistent performance on a new domain \cite{volpi2018generalizing, carlucci2019domain}, due to 
the cross-domain distributional shift
\cite{nado2020evaluating}. To address the potential discrepancies between the training and test domains, 
a number of works \cite{huang2020self, chattopadhyay2020learning, zhou2020learning} have been proposed to learn domain-invariant features using the training data from multiple source domains \cite{seo2019learning, li2017deeper} to improve the model's generalization capability across domains. However, acquiring multi-domain training data is both challenging and expensive. Alternatively, a more practical but 
less investigated solution is to train the model on a single source domain data and enhance its capability of generalizing to other unseen domains (see the example in Fig.~\ref{fig:sdg_pacs}). This emerging learning paradigm is referred to as {\it single domain generalization} \cite{qiao2020learning}. 

Existing works on {\it single domain generalization} \cite{volpi2018generalizing,volpi2019addressing, qiao2020learning,zhao2020maximum,huang2020self} try to improve the generalization capability through adversarial domain augmentation (ADA), which synthesizes new training images in an adversarial way
to mimic virtual challenging domains. The model therefore learns the domain-invariant features to improve its generalization performance. In this work, we propose 
to tackle the single domain generalization challenge from a different perspective, building an adaptive normalization in the ADA framework to improve the model's domain generalization capability. The motivation behind this idea is that the batch normalization (BN), used by the existing works on single domain generalization, lacks the domain generalization capability due to the discrepancy between the training and testing data statistics. More specifically, at the training stage, BN standardizes the feature maps by the statistics estimated on a batch of training data. The exponential moving averages (EMA) of the training statistics are then applied during testing and make the computation graph inconsistent between the training and testing. In {\it single domain generalization}, the testing domain statistics are usually different from the training domain statistics. Therefore, applying the statistics estimated from the training to testing will likely result in performance drops. BN-Test \cite{nado2020evaluating} has been proposed to substitute the EMA of the training statistics with testing batch statistics for remedy. However, this would require batching test data and the testing performance becomes dependent on the testing batch size.

\begin{figure}[tp!]
\centering
\includegraphics[height=5.3cm]{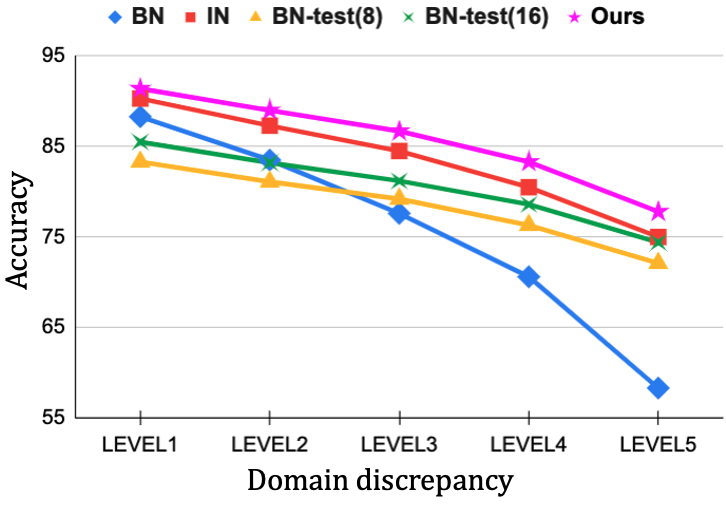} \vspace{-2.5mm}
\caption{\small Accuracy of five different normalization methods for single domain generalization on CIFAR-10-C, compared at five different levels of domain discrepancy brought by corruptions. Methods include BN, IN, BN-Test (different batch sizes), and our method. 
See detailed experimental settings in Sec.~\ref{sec:exp}.}
\label{fig:cifar_bn} \vspace{-5.5mm}
\end{figure}

Fig.~\ref{fig:cifar_bn} verifies the deficiency of BN by comparing five different normalizations for single domain generalization on CIFAR-10-C \cite{hendrycks2018benchmarking}. Due to the domain discrepancy between the training and testing data, BN underperforms instance normalization (IN) \cite{ulyanov2016instance}. The performance gap increases significantly with the level of the domain discrepancy getting higher. Although BN-Test(16) improves the performance over BN when the domain discrepancy is large, it requires a batch size of 16 during inference which might not be available in practice. Additionally, as shown in Table~\ref{tab:mnist_ab} in Appendix, BN-Test might not work on some other benchmarks even with a large batch size. This motivating example shows that using BN by default is sub-optimal and inspires us to explore the normalization algorithm to improve the model's generalization capability across domains. 

To this end, we propose a novel adaptive form of normalization named as {\it adaptive standardization and rescaling normalization} ({\ours}), in which the standardization and rescaling statistics are both learned to be adaptive to each individual input sample. When being used with ADA \cite{volpi2018generalizing,volpi2019addressing,huang2020self}, {\ours} can learn the normalization statistics by approximately optimizing a robust objective, making the statistics be adaptive to the data coming from different domains, and hence helping the model to generalize better across domains than traditional normalization approaches. We also show that {\ours} can be viewed as a generic form of the traditional normalization approaches including BN, IN, layer normalization (LN) \cite{ba2016layer}, group normalization (GN) \cite{wu2018group}, and switchable normalization (SN) \cite{luo2018differentiable}.


Our main contributions are as follows: (1) We propose a novel adaptive normalization, the missing ingredient for current works on ADA for single domain generalization. To the best of our knowledge, the proposed ASR-Norm is the first to learn both standardization and rescaling statistics in normalization with neural networks. 
(2) We show that ASR-Norm can bring consistent improvements to the state-of-the-art ADA approaches on three commonly used single domain generalization benchmarks. The performance gain is agnostic to the choice of ADA methods and becomes more significant as the domain discrepancy increases.

\section{Related work}
\noindent{\bf Domain Discrepancy.} 
{\it Domain adaptation} \cite{ganin2015unsupervised,tzeng2017adversarial, saenko2010adapting,zhao2018adversarial} alleviates the effect of domain discrepancy by allowing the model to see unlabeled target domain data during training. By contrast, {\it domain generalization} aims to learn domain-invariant representations so as to improve the generalization without any access to the target domains. The majorities of the literatures on {\it domain generalization} \cite{qiao2020learning,huang2020self,chattopadhyay2020learning, seo2019learning,zhou2020learning} focus on learning the domain invariant knowledge from multiple source domains. Another line of works studies a more challenging and realistic setting, {\it single domain generalization}, which is the focus of this paper. A few recent works on single domain generalization \cite{volpi2018generalizing,volpi2019addressing,qiao2020learning,zhao2020maximum} show that ADA can effectively improve the generalization and robustness of models by synthesizing virtual images during training. 
For example, \citet{volpi2018generalizing} generated the virtual images with adversarial updates on the input images while maintaining the semantic similarities with the original images. \citet{volpi2019addressing} proposed a random search data augmentation ({RSDA}) algrorithm
that picks one set of the most challenging transformations in terms of inference accuracy to augment the image for training.
On the other side, \citet{huang2020self} proposed a representation self-challenging (RSC) algorithm to virtually augment challenging data by shutting down the dominant neurons that have the largest gradients during training. 
Our work builds an adaptive normalization scheme and combines it with ADA which yields a better generalization ability across domains than the state-of-the-art single domain generalization approaches.\\

\noindent{\bf Robustness against Distributional Shifts.} 
{\it Adversarial training} \cite{goodfellow2014explaining,madry2018towards} aims to make models be robust to adversarial examples with {\it imperceptible} perturbations added to the inputs. However, our work focuses on the generalization ability to more {\it perceptible} and natural distributional shifts brought by the domain discrepancy. Meanwhile, there are also works trying to improve model robustness to natural and perceptible noises with pretrain \cite{hendrycks2020many,hendrycks2019using,hendrycks2020pretrained}, data augmentation \cite{hendrycks2020many,hendrycks2019augmix,qiao2020learning}, contrastive learning \cite{chen2020big}, stochastic networks \cite{fan2020bayesian,fan2021contextual}, etc. This setting can be included into the single domain generalization framework by viewing different distortions as different domains. Our work is compatible with these methods and can potentially further improve the performance in single domain generalization.\\

\noindent{\bf Normalization in Neural Networks.} 
Since the invention of 
BN \cite{ioffe2015batch},
various normalization techniques \cite{nado2020evaluating,seo2019learning,jia2019instance,luo2018differentiable,wu2018group,li2020feature,wang2019transferable,salimans2016improved,xu2019understanding,ba2016layer,ulyanov2016instance} have been proposed for different applications.
Batch-instance normalization (BIN) \cite{nam2018batch} and SN \cite{luo2018differentiable,shao2019ssn}
were proposed to generalize the fixed normalizations by combining their standardization statistics with the learnable weights. However, their restrictive forms still do not allow enough adaptivity. More recently, instance-level meta (ILM) normalization \cite{jia2019instance} focused on improving the rescaling performance by using neural networks to learn the rescaling statistics.
Before our work, the majorities of the works on normalization do not study the generalization ability under domain shift. Below are a few exceptions: BN-Test \cite{nado2020evaluating} uses testing time statistics, 
making its performance highly depend on the testing batch size; DSON \cite{seo2019learning} requires multi-source training data to learn separate BIN for each domain and ensemble the normalization during testing. Compared with these methods, 
our normalization scheme does not impose dependencies between the testing samples, requires only the single domain training data, and is generic by learning both the standardization and rescaling statistics.

\section{Adversarially Adaptive Normalization}
\begin{figure*}[tp!]
\centering
\includegraphics[width=17.3cm]{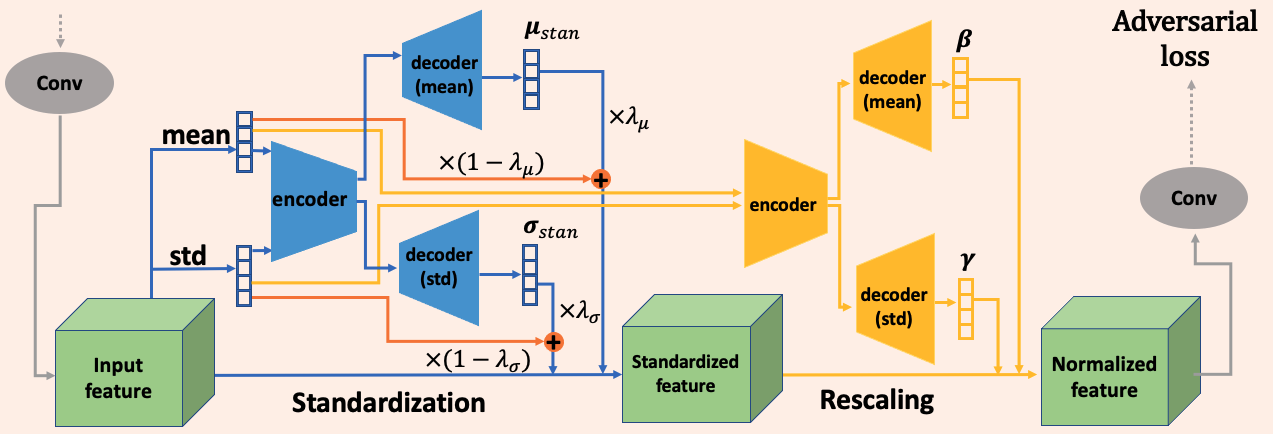}\vspace{-0.5mm}
\caption{\small Illustration of the {\ours} layer. It can be divided into two steps: standardization and rescaling. We use encoder-decoder structured networks to learn both the standardization and rescaling statistics from the channel-wise statistics of the input. For standardization, we combine the learned statistics and native statistics with adaptive weights for stabilizing the training process.}\vspace{-2.5mm}
\label{fig:lslr_block} 
\end{figure*}

\subsection{Background}
\subsubsection{Single Domain Generalization Problem Setup}
Consider a supervised learning problem with the training dataset $\mathcal{D}_s \sim P_s$ , where $P_s$ denotes the source domain distribution. The goal of single domain generalization is to train a model using single source domain data $\mathcal{D}_s$ to correctly classify the images from a target domain distribution $P_t$ unavailable during training. In this case, using the vanilla empirical risk minimization (ERM) \cite{vapnik2013nature} solely on the source domain ${P}_s$
as the training objective could be sub-optimal and yield a model that does not generalize well to unseen domains \cite{volpi2018generalizing}. Denote $\theta$ as the model parameters and $l: \mathcal{X}\times \mathcal{Y}\rightarrow \mathbb{R}$ as the loss function. To help the model better generalize to the unseen domains, a robust objective that considers a worst-case problem around the source domain $P_s$ has been proposed in \cite{sinha2018certifying} as
\vspace{-1mm}
\begin{equation}
    \mathcal{L}_{R}(\theta):= \sup_{P: D(P, P_s)\leq \rho}\E _{\{X,Y\}\sim P}[l(\theta; X, Y)],\vspace{-1mm}
\end{equation}
where $D(P,Q)$ is a distance metric on domain distributions. The robust objective $\mathcal{L}_{R}$ enables training a model which performs well on the distributions that are $\rho$-distance away from the source domain distribution $P_s$. However, it is generally difficult to directly optimize the robust objective. 

\subsubsection{Adversarial Domain Augmentation}\label{sec:ada}
The optimization of $\mathcal{L}_R$ can be converted to a Lagrangian optimization problem with a penalty parameter $\eta$ and solved via a min-max approach: 
\vspace{-1mm}
\begin{equation}
    \mathcal{L}_{RL}:=\sup_P \{\E_P[l(\theta;(X,Y))] - \eta D (P, P_s)\}.\label{eq:obj_rl}\vspace{-1mm}
\end{equation}
Defining the distance $D$ between two distributions by the Wasserstein distance \cite{volpi2018generalizing} on a learned semantic space, the gradient of $\mathcal{L}_{RL}$, under suitable condition, can then be written as
$\nabla_\theta L_{RL} =\E _{(X,Y)\sim P_s}[\nabla_\theta l(\theta; (X_\eta^*, Y))],$
where $ X_\eta^* :=\argmax_{x\in \mathcal{X}}\{l(\theta; (x,Y)) - \eta c_\theta ((x,Y)), (X,Y))\},$ and $c_\theta$ is a learned distance measure over the space $\mathcal{X}\times \mathcal{Y}$ (see details in Appendix Sec.~\ref{sec:app_ada}) \cite{volpi2018generalizing,boyd2004convex}.
Gradient ascent was proposed to find approximation $ X_\text{aug}$, to $ X_\eta^*$, which maximizes the prediction loss $l$ while maintaining close semantic distance to the original image $X$. The synthesized image, $ X_\text{aug}$, with its label is then appended to the training dataset. This phase, referred to as the maximization phase,  is alternated with the minimization phase, where we optimize $\theta$ to minimize the prediction loss on both the original and augmented data. 

\subsubsection{Standardization and Rescaling in Normalization}
This section briefly introduces the normalization framework. Denote the input tensor of a normalization layer by $\xv\in \mathbb{R}^{C\times H\times W}$, where $C$ denotes the number of channels of the tensor, $H$ denotes the height, and $W$ denotes the width.
A typical normalization layer consists of two steps: standardization and rescaling. During standardization, the mean and standard deviation, $\muv_\text{stan}, \sigmav_\text{stan}\in \mathbb{R}^C$, are derived from the input tensors and used to standardize the input tensors (a small positive number, $\epsilon$, is added to $\sigmav_\text{stan}$ to avoid numerical issues). At the rescaling step, the standardized activations $\xv_\text{stan}$ are rescaled with the rescaling statistics $\gammav, \betav\in \mathbb{R}^C$:
\vspace{-1mm}
\begin{equation}
    \begin{cases}
     \xv_\text{stan} = (\xv - \muv_\text{stan}) / (\sigmav_\text{stan} + \epsilon),\\
      \xv_\text{norm} = \xv_\text{stan} * \gammav  + \betav.
    \end{cases}\vspace{-1mm}
\end{equation}


Different types of normalization share the similar formula but differ in the way of measuring the statistics. For example, BN computes the statistics for each batch, while IN, GN and LN compute the statistics for each sample with different channel groups. SN combines the statistics of these normalizations with learnable weights. In this work, we aim to build an adaptive normalization with statistics learned by neural networks to improve the model's domain generalization capability.





\subsection{ASR-Norm: Adaptive Standardization and Rescaling Normalization}
In this section, we propose a novel generic and adaptive normalization technique, {\bf \ours}, short for {\bf A}daptive {\bf S}tandardization and {\bf R}escaling {\bf Norm}alization. {\ours} learns both standardization and rescaling statistics with auto-encoder structured neural networks.
{\ours} is adaptive to the individual input sample and therefore has a consistent computational graph between training and testing. Additionally, we introduce a residual term for the standardization statistics to stabilize the learning process. Using {\ours} with ADA can learn robust normalization statistics to enhance the model's domain generalization ability. An overview of the proposed {\ours} is shown in Fig.~\ref{fig:lslr_block} with details in the following sections.


\subsubsection{Adaptive Standardization (AS)}
{\bf A Functional Form.} 
The standardization statistics can be viewed as the functions of the input tensor $\xv$, i.e.,
\vspace{-1mm}
\begin{equation}
    \muv_\text{stan} = f(\xv), \sigmav_\text{stan} = g(\xv).\vspace{-1mm}
\end{equation}

\begin{remark}
This form generalizes a variety of normalization methods:
\vspace{-3mm}
\begin{itemize}
    \item If $f,g$ are constant functions with values being the statistics $\muv_0, \sigmav_0$ computed from the whole training set, this form is equivalent to {BN} with all training data as a batch: $f_\text{BN}(\xv) = \muv_0, g_\text{BN}(\xv) = \sigmav_0;$ \vspace{-3mm}
    \item If $f,g$ are domain-wise constant functions with values being the statistics $\muv_d, \sigmav_d$ computed for each domain $P_d$ in the whole training set, this form is equivalent to {domain-specific BN} (DSBN) with each domain data as a batch: $f_\text{DSBN}(\xv) = \sum_{d=1}^{N_d} \muv_d {\bf 1}_{\{\xv\in P_d\}},  g_\text{DSBN}(\xv)= \sum_{d=1}^{N_d} \sigmav_d {\bf 1}_{\{\xv\in P_d\}},$ where $N_d$ is the number of domains;\vspace{-3mm}
    \item If $f,g$ are the mean and std functions for each group of channels respectively, this form is equivalent to {GN}, which also generalizes {IN} and {LN}; \vspace{-3mm}
    \item If $f,g$ are weighted combinations of BN, IN, and LN, this form is equivalent to {SN}:\vspace{-1mm}
    
    $f_\text{SN}(\xv) = w_1 f_\text{BN}(\xv) + w_2 f_\text{IN}(\xv) + w_3 f_\text{LN}(\xv)$,\vspace{-1mm}
    
    $g_\text{SN}(\xv)= w_1' g_\text{BN}(\xv) + w_2' g_\text{IN}(\xv)+ w_3' g_\text{LN}(\xv).$\vspace{-2mm}
\end{itemize}



\end{remark}


From the above observations, we notice that each normalization imposes a rather restrictive form of $f$ and $g$, which limits the flexibility of the normalization layers. As a result, we propose to use neural networks to fully learn the functions $f,g$ so as to provide a generic way to obtain standardization statistics.

{\bf Standardization Neural Networks.} Our goal is to construct mappings from the feature map $\xv$ to the learned statistics $\muv_\text{stan}$ and $\sigmav_\text{stan}$. To lower the computational cost and leverage the original statistics information contained in the convolutional feature maps, the standardization network chooses the channel-wise mean and standard deviation statistics, instead of the original feature map, as the input to learn the standardization statistics.

Formally, denote channel-wise mean and standard deviation statistics vector of $\xv$ by $\muv, \sigmav \in \mathbb{R}^C$, respectively. The per-channel mean and standard deviation are expressed as
\vspace{-1mm}
\begin{equation}
    \begin{cases}
      \mu_c = \sum_{i=1}^H\sum_{j=1}^W \xv_{cij} / (H\times W),\\
       \sigma_c = \sqrt{\sum_{i=1}^H\sum_{j=1}^W (\xv_{cij} - \mu_c)^2 / (H\times W)}\label{eq:stan}
    \end{cases}\vspace{-1mm}
\end{equation}
for $ c=1,..., C$.

We make use of an encoder-decoder structure \cite{hu2018squeeze,jia2019instance} to learn $\muv_\text{stan}$ and $\sigmav_\text{stan}$ from $\muv$ and $\sigmav$ respectively, where the encoder extracts global information by interacting the information of all channels and the decoder learns to decompose the information for each channel. For the sake of efficiency, both the encoder and decoder consist of one fully-connected layer, forming a bottleneck connected by a non-linear activation function, ReLU.  Another ReLU \cite{nair2010rectified} layer is applied to the output of $\sigmav_\text{stan}$ to make sure it is non-negative: 
\vspace{-1mm}
\begin{equation}
    \begin{cases}
     \muv_\text{stan} = f(\xv) := f_\text{dec}(\text{ReLU}(f_\text{enc}(\muv))),\\
      \sigmav_\text{stan} = g(\xv) := \text{ReLU}(g_\text{dec}(\text{ReLU}(g_\text{enc}(\sigmav)))),\label{eq:learn_stan}
    \end{cases}\vspace{-1mm}
\end{equation}
where $f_\text{enc}, f_\text{dec}, g_\text{enc}, g_\text{dec}$ are fully-connected layers. The encoders project the input to the hidden space $\mathbb{R}^{C_\text{stan}}$, while the decoders project it back to the space $\mathbb{R}^{C}$, where $C_\text{stan} < C$. In practice, we find that sharing the encoders for $\muv_\text{stan}$ and $\sigmav_\text{stan}$ would not affect the performance and save memory. Therefore, we let $f_\text{enc}= g_\text{enc}$.

{\bf Residual Learning.} In the early training stage, the learning process of the standardization networks can be unstable and lead to numerical issues when the learned statistics are inaccurate. For example, $\sigmav_\text{stan}$ could be very small due to the ReLU activation, making the scale of $\xv_\text{stan}$ very large. One remedy for this is to impose additional restrictive activation functions, such as sigmoid, on $\sigmav_\text{stan}$ so that the output values $\sigmav_\text{stan}$ are bounded. However, this could harm the training process for the case that the scale of $\xv_\text{stan}$ is large when the scale of $\xv$ is large. 
To retain the flexibility of $\sigmav_\text{stan}$'s scale, we stick to the ReLU activation and solve the numerical issue by introducing a residual term for regularization. In particular, we make both $\muv_\text{stan}$ and $\sigmav_\text{stan}$ as weighted sums of the learned and original statistics:
\vspace{-1mm}
\begin{equation}
    \begin{cases}
     \muv_\text{stan} = \lambda_\muv f_\text{dec}(\text{ReLU}(f_\text{enc}(\muv))) + (1- \lambda_\muv) \muv,\\
      \sigmav_\text{stan} = \lambda_\sigmav \text{ReLU}(g_\text{dec}(\text{ReLU}(g_\text{enc}(\sigmav))) + (1- \lambda_\sigmav) \sigmav.
    \end{cases}\vspace{-1mm}
\end{equation}
The weights $\lambda_\muv, \lambda_\sigmav$ are learnable parameters ranging from $0$ to $1$ (bounded by sigmoid).
They are both initialized with small values close to $0$ so that the standardization process is stable in the early stage and the network can gradually switch to the learned statistics in the later stage (see Fig.~\ref{fig:weight_adapt_only}).

\begin{remark}
All statistics and transformation operations in AS are computed for each sample independently. Unlike BN, AS removes the dependencies among samples during standardization. The computational graph of AS is consistent between the training and testing.
\end{remark}

\subsubsection{Adaptive Rescaling (AR)}
{\bf Data-dependent rescaling parameters.} Most normalization layers use the learnable parameters $\gammav, \betav \in \mathbb{R}^C$ to rescale standardized output $\xv_\text{stan}$, resulting in a rescaling process that is uniform to all samples. Recently, \citet{jia2019instance} found that making the rescaling parameters dependent on the samples and allowing different samples to have different rescaling parameters can bring performance gain to instance-level normalization for a variety of in-domain tasks. Inspired by this observation, we construct a rescaling network for the adaptive rescaling in {\ours}. 

{\bf Rescaling Neural Networks.}
Similar to the standardization networks, we define the following network to learn the rescaling parameters from the original statistics $\muv, \sigmav$:
\vspace{-1mm}
\begin{equation}
    \begin{cases}
     \betav = \psi(\xv) := \text{tanh}(\psi_\text{dec}(\text{ReLU}(\psi_\text{enc}(\muv)))) + \betav_\text{bias},\\
      \gammav = \phi(\xv) := \text{sigmoid}(\phi_\text{dec}(\text{ReLU}(\phi_\text{enc}(\sigmav)))) + \gammav_\text{bias},
    \end{cases}\vspace{-1mm}
\end{equation}
where $\phi_\text{enc}$, $\phi_\text{dec}$, $\psi_\text{enc}$, $\psi_\text{dec}$ are fully-connected layers, and $\text{sigmoid(), tanh()}$ are activation functions to ensure the rescaling statistics are bounded. The encoders project the inputs to the hidden space $\mathbb{R}^{C_\text{rescale}}$ with $C_\text{rescale} < C$. The decoders project the encoded feature back to the space $\mathbb{R}^{C}$. $\gammav_\text{bias}, \betav_\text{bias}\in \mathbb{R}^C$ are learned parameters and are initialized with ones and zeros, respectively, as the traditional rescaling parameters. The encoders for $\betav, \gammav$ are also shared according to the same reason as the standardization networks, i.e., $\phi_\text{enc}= \psi_\text{enc}$.


\subsubsection{Adversarially Adaptive Normalization}
The parameters in {\ours} can be learned together with the model $\theta$ by the robust objective as $\theta$ in Eq~\ref{eq:obj_rl}. With ADA, the objective can be optimized approximately, and {\ours} can learn the normalization statistics to be adaptive to the data coming from different domains, thus helping the model generalize well across domains.
Note that the proposed normalization is agnostic to the choice of ADA. In the experiments, we show the implementations and performances of our method coupling with three different ADA methods.

\section{Experiments}\label{sec:exp}

\subsection{Experimental Settings}

{\bf Datasets.} 
We conduct experiments on  three standard benchmarks for single domain generalization \cite{volpi2018generalizing,volpi2019addressing,zhao2020maximum,qiao2020learning}, including {\it Digits, CIFAR-10-C, }and {\it PACS}. 

(1) {\it Digits:} This benchmark consists of five digits datasets: MNIST \cite{lecun1989backpropagation}, SVHN \cite{netzer2011reading}, MNIST-M \cite{ganin2015unsupervised}, SYN \cite{ganin2015unsupervised}, and USPS \cite{hull1994database} (see examples in Fig.~\ref{fig:sdg} in Appendix). We use MNIST as the source domain, and the rest as the target domains. All images are resized to $32\times 32$ pixels and the channels of MNIST and USPS are duplicated so that different datasets have compatible shapes.

\begin{figure}[htp!]
\centering
\includegraphics[height=2.7cm]{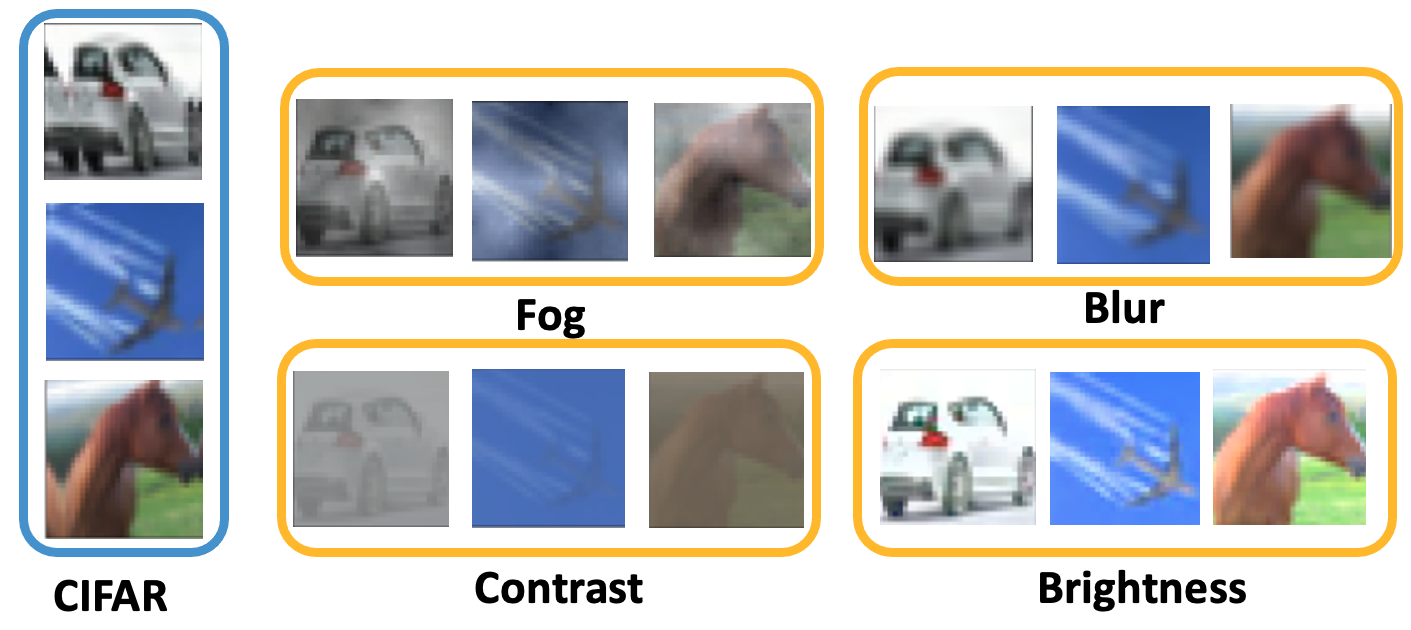}\vspace{-2.5mm}
\caption{\small Example images  with level $5$ corruption from $4$ corruption types in the CIFAR-10-C benchmark.}
\label{fig:cifar} \vspace{-3.5mm}
\end{figure}

(2) {\it CIFAR-10-C} \cite{hendrycks2018benchmarking}:  This benchmark is proposed to evaluate the robustness to $19$ types of corruptions with $5$ levels of intensities. The original CIFAR-10 \citep{krizhevsky2009learning} is used for training and corruptions are only applied to the testing images (see examples in Fig.~\ref{fig:cifar}). The level of domain discrepancy can be measured by the corruption intensity.

(3) {\it PACS} \cite{li2017deeper}: A recent benchmark for domain generalization containing four domains: \emph{art paint}, \emph{cartoon}, \emph{sketch}, and \emph{photo}
(see examples in Fig.~\ref{fig:sdg_pacs}). This dataset is considered to pose a challenging distributional shift for domain generalization. For this dataset, we consider two settings: 1) training a model with one domain data and testing with the rest three domains; 2) training a model on three domains and test on the remaining one. The second setting is widely used for multi-source domain generalization with the domain label available for training. In our experiment for single domain generalization, we remove the domain label and mix data from multiple source domains during training.



{\bf Implementation Details. } For \ours, unless otherwise noted,
$C_\text{stan},C_\text{rescale}$ are set to $C/2, C/16$, respectively, and $\lambda_\muv,\lambda_\sigmav$ are both initialized as \text{sigmoid}(-3).

(1) For {\it Digits}, we use the ConvNet architecture \cite{lecun1989backpropagation} ({\it conv-pool-conv-pool-fc-fc-softmax}) with ReLU following each convolution. Since this model does not have any normalization layer, {\ours} is inserted after each conv layer before ReLU. We adopt {RSDA} \cite{volpi2019addressing} for domain augmentation, which is the state-of-the-art (SOTA) on this benchmark, and follow its experimental settings: setting the size of transformation tuples as $3$, using a subset of $1,000$ samples from training set to select the worst transformation, conducting the search every $1,000$ steps, running a total of $10^6$ weight updates, and using Adam \cite{kingma2014adam} optimizer with learning rate $10^{-4}$ and mini-batch size $32$. 

(2) For {\it CIFAR-10-C}, we use the Wide Residual Network (WRN) \cite{zagoruyko2016wide} with $40$ layers and widen factor $4$. We follow the experiment setting in \cite{zagoruyko2016wide} to use SGD with Nesterov momentum ($0.9$) and batch size $128$. The learning rate starts at $0.1$ with cosine annealing \cite{loshchilov2016sgdr}. Models are trained for $200$ epochs in total. We replace the default BN with {\ours} and train with {ADA} \cite{volpi2018generalizing}, where the augmentation is performed for every $1000$ steps of training with three augmentations in total, and each augmentation step consists of $25$ gradient updates. 

(3) For {\it PACS}, we use a ResNet-18 pretrained on ImageNet as the backbone and add a fully-connected layer for classification as the setting in \cite{huang2020self} which presented the SOTA domain generalization performance on PACS. The BN in the ResNet-18 model is substituted by \ours. We find that directly replacing BN with {\ours} during the fine tuning stage fails to work. The main reason is that the models are sensitive to the scale of the activations after pretrain and replacing the rescaling parameters with the AR networks could lead to significant changes in the activation scales. To remedy the issue, we resume $\betav_\text{bias}$ and $\gammav_\text{bias}$ from the BN layers in the pretrained model and add learnable weights $\lambda_\betav, \lambda_\gammav$ to the first terms:
\vspace{-1mm}
\begin{equation}
    \begin{cases}
    \betav = \psi(\xv) = \lambda_\betav \text{tanh}(\psi_\text{dec}(\text{ReLU}(\psi_\text{enc}(\muv)))) + \betav_\text{bias},\\
      \gammav = \phi(\xv) = \lambda_\gammav \text{sigmoid}(\phi_\text{dec}(\text{ReLU}(\phi_\text{enc}(\sigmav)))) + \gammav_\text{bias},
    \end{cases}\vspace{-1mm}
\end{equation}
where we initialize $\lambda_\betav$ and $\lambda_\gammav$ with a small value, $\text{sigmoid}(-5)$, to smooth the learning process. In this way, the scale of $\xv_\text{norm}$ will be close to that of the pretrained models initially. The network can gradually learn to make use of the learned statistics. During training, the model learns {\ours} with the RSC procedure \cite{huang2020self} by SGD optimizer with the initial learning rate as $0.004$, which decays to $10\%$ after 24 epochs. Models are trained for $30$ epochs with the batch size of $128$.


\subsection{Main Results: Comparing with the SOTA }
Our approach improves upon the SOTA single domain generalization results by $1.6\%$, $2.7\%$, and $6.3\%$ on the Digits, CIFAR-10-C, and PACS benchmarks, respectively. 

\begin{table}[ht!] 
\vspace{-2mm}
\centering
\resizebox{0.95\columnwidth}{!}{
\begin{tabular}{@{}lccccclllllll@{}}\toprule
Method & SVHN & MNIST-M & SYN & USPS & Avg.\\ \midrule
ERM & 27.8 & 52.7 & 39.7  & 76.9  & 49.3 \\
CCSA\cite{motiian2017unified} & 25.9 & 49.3 & 37.3 & 83.7 & 49.1 \\
d-SNE\cite{xu2019d} & 26.2 & 51.0 & 37.8 & {\bf 93.2} & 52.1 \\
JiGen\cite{carlucci2019domain} & 33.8 & 57.8 & 43.8 & 77.2 & 53.1 \\
ADA\cite{volpi2018generalizing} & 35.5 & 60.4 & 45.3 & 77.3 & 54.6 \\
M-ADA\cite{qiao2020learning} & 42.6 & 67.9 & 49.0 & 78.5 & 59.5 \\
ME-ADA\cite{zhao2020maximum} & 42.6 & 63.3 & 50.4 & 81.0 & 59.3 \\
RSDA\cite{volpi2019addressing} & 47.4$\pm$4.8 & {81.5}$\pm$1.6 & 62.0$\pm$1.2 & 83.1$\pm$1.2 &68.5\\  \midrule
RSDA+ASR (Ours) & {\bf 52.8$\pm$3.8} & {80.8}$\pm$0.6 & {\bf 64.5$\pm$1.1} & 82.4$\pm$1.4& {\bf 70.1} \\ 
$\Delta$ to RSDA & 5.4 & -0.7 & { 2.5} & -0.7 & 1.6 \\
\bottomrule
\end{tabular}}\vspace{-1mm}\caption{Single domain generalization accuracies on Digits. MNIST is used as the training set, and the results on different testing domains are reported in different columns. }\label{tab:mnist}\vspace{-2mm}
\end{table}

{\bf Results on Digits. }Table~\ref{tab:mnist} shows the results on the {\it Digits} benchmark. The proposed method, {\ours}, is combined with RSDA and compared with methods including: (1) ERM which uses cross-entropy loss for training, without domain augmentation; (2) CCSA \cite{motiian2017unified} which regularizes latent features to improve generalization; (3) d-SNE \cite{xu2019d} which uses stochastic neighborhood embedding techniques and a novel modified-Hausdorff distance for training; (4) JiGen \cite{carlucci2019domain} which adds the patch order prediction as auxiliary; (5) ADA \cite{volpi2018generalizing}; (6) M-ADA \cite{volpi2018generalizing} which uses Wasserstein auto-encoder and meta-learning to improve ADA; (7) ME-ADA \cite{zhao2020maximum} which adds entropy regularization to help ADA; (8) RSDA \cite{volpi2019addressing} which does not use normalization. Our method outperforms both the baseline and the SOTA methods on average. In particular, our method improves the performance on challenging domains like SVHN and SYN. We have similar observations on USPS as \cite{qiao2020learning}, where ADA-based methods are not as good as CCSA or d-SNE due to USPS's strong similarity with MNIST (see Fig.~\ref{fig:sdg}).


\begin{table}[ht] 
\vspace{-2mm}
\centering
\resizebox{0.95\columnwidth}{!}{
\begin{tabular}{@{}lccccccllllllll@{}}\toprule
Method & Level 1 & Level 2 & Level 3 & Level 4 & Level 5 & Avg.\\ \midrule
ERM & 87.8$\pm$0.1 & 81.5$\pm$0.2 & 75.5$\pm$0.4 & 68.2$\pm$0.6 & 56.1$\pm$0.8 & 73.8\\
PGD\cite{madry2017towards} & 73.1$\pm$0.5 & 69.0$\pm$1.1 & 64.1$\pm$1.3 & 58.0$\pm$2.2 & 48.9$\pm$2.8 & 61.6 \\
ST \cite{hendrycks2019using} & - & - & - & - & - & 76.9\\
TTT \cite{sun2020test} & - & - & - & - & - & 84.4\\
ADA\cite{volpi2018generalizing} & 88.3$\pm$0.6 & 83.5$\pm$2.0 & 77.6$\pm$2.2 & 70.6$\pm$2.3 & 58.3$\pm$2.5 & 75.7\\
M-ADA\cite{qiao2020learning} & 90.5$\pm$0.3 & 86.8$\pm$0.4 & 82.5$\pm$0.6 & 76.4$\pm$0.9 & 65.6$\pm$1.2  &80.4\\
ME-ADA\cite{zhao2020maximum} & - & - & - & - & - & 83.3\\
\midrule
ERM+ASR & 89.4$\pm$0.2 & 86.1$\pm$0.2 & 82.9$\pm$0.3 & 78.6$\pm$0.6 & 72.9$\pm$1.0&82.0  \\
ADA+ASR (Ours) & {\bf 91.5$\pm$0.2} & {\bf 89.3$\pm$0.6} & {\bf 86.9$\pm$0.5} & {\bf 83.7$\pm$0.7} & {\bf 78.4$\pm$0.8} &{\bf 86.0}\\
$\Delta$ to ADA & 3.2 & 5.8 & 9.3 & 13.1 & 20.1 & { 10.3}\\
\bottomrule
\end{tabular}}\vspace{-1mm}\caption{Single domain generalization accuracies on CIFAR-10. CIFAR-10 is used as the training domain, while CIFAR-10-C with different corruption types and corruption levels are used as the testing domains. }\label{tab:cifar}\vspace{-2mm}
\end{table}

{\bf Results on CIFAR-10-C.} We report the average accuracies of the $19$ corruption types for each level of intensity 
on CIFAR-10-C in Table~\ref{tab:cifar} (we also provide results for each corruption type in Fig.~\ref{fig:cifar_type} in Appendix Sec.~\ref{sec:app_results}). ERM, ADA\cite{volpi2018generalizing}, M-ADA\cite{volpi2019addressing}, ME-ADA\cite{zhao2020maximum}, ST\cite{hendrycks2019using}, and TTT\cite{sun2020test} are used for the comparison on this benchmark. We also include the results by using project gradient descent (PGD) \cite{madry2017towards} for adversarial training. The results show {\ours} makes significant improvement over ADA and all other improved ADA-based methods such as M-ADA and ME-ADA, as well as self-supervised method ST and test-time training method TTT. Similar to the results on the Digits benchmark, {\ours} achieves larger improvements on the more challenging domains, i.e., domains with more intense corruptions, than on the less challenging ones. We also note that the standard PGD training for defending against adversarial examples does not help improve robustness to perceivable noises in this dataset. 


\begin{table}[ht] 
\vspace{-2mm}
\centering
\resizebox{0.95\columnwidth}{!}{
\begin{tabular}{@{}lccccclllllll@{}}\toprule
Method & Artpaint & Cartoon & Sketch & Photo & Avg.\\ \midrule
ERM & 70.9 & 76.5 & 53.1 & 42.2 & 60.7\\
RSC\cite{huang2020self} & 73.4 & 75.9 & 56.2 & 41.6 & 61.8\\ \midrule
RSC+ASR (Ours) & {\bf 76.7} & {\bf 79.3} & {\bf 61.6} & {\bf 54.6} & {\bf 68.1}\\ 
$\Delta$ to RSC  & 3.3 & 3.4 & 5.4 & 13.0 & 6.3\\
\bottomrule
\end{tabular}}\vspace{-1mm}\caption{Single domain generalization accuracies on PACS. One domain (name in column) is used as the training set and the other domains are used as the testing set. }\label{tab:pacs}\vspace{-2mm}
\end{table}

{\bf Results on PACS.} In Table~\ref{tab:pacs}, we show the results on PACS where we use one domain for training and the rest three for testing. The reported numbers are the average accuracies across the testing domains. Again, {\ours} improves the performance of RSC significantly, especially on challenging domains. Table~\ref{tab:pacs_multi} shows the results on PACS for the multi-source domain setting, where we do not utilize domain labels during training. Similarly, {\ours} outperforms not only the SOTA performance reported by RSC but also the other SOTA methods, including those that make use of the domain labels, such as DSON and MetaReg.

\begin{table}[ht] 
\vspace{-2mm}
\centering
\resizebox{0.99\columnwidth}{!}{
\begin{tabular}{@{}lccccclllll@{}}\toprule
Method & Artpaint & Cartoon & Sketch & Photo & Avg.\\ \midrule
ERM & 79.0 & 73.9 & 70.6 & {\bf 96.3} & 79.9\\
JiGen\cite{carlucci2019domain} & 79.4 & 75.3 & 71.4 & 96.0 & 80.5\\
MetaReg \cite{balaji2018metareg} & 83.7 & 77.2 & 70.3 & 95.5 & 81.7\\
Cutout\cite{devries2017improved} & 79.6 & 75.4 & 71.6 & 95.9 & 80.6\\
DropBlock\cite{ghiasi2018dropblock} & 80.3 & 77.5 & 76.4 & 95.6 & 82.5\\
AdversarialDropout\cite{park2017adversarial} & 82.4 & 78.2 & 75.9 & 91.1 & 83.1\\
BIN\cite{seo2019learning} & 82.1 & 74.1 & 80.0 & 95.0 & 82.8\\
SN\cite{luo2018differentiable} & 78.6 & 75.2 & 77.4 & 91.1 & 80.6\\
DSON\cite{seo2019learning} & 84.7 & 77.7 & { 82.2} & 95.9 & 85.1\\
RSC\cite{huang2020self} & 83.4 & 80.3 & 80.9 & 96.0 & 85.2\\ \midrule
RSC+ASR (Ours) & {\bf 84.8} & {\bf 81.8} & {\bf 82.6} &  96.1 & {\bf 86.3}\\
$\Delta$ to RSC & 1.4 & 1.5 & 1.7 & 0.1 & 1.1\\ 
\bottomrule
\end{tabular}}\vspace{-1mm}\caption{Domain generalization accuracies on PACS. One domain (name in column) is used as the test set and the other domains are used as the training sets. During training, no domain identification is used. 
}\label{tab:pacs_multi}\vspace{-5mm}
\end{table}

\subsection{Result Analysis}
{\bf On the Effect of Normalization.} In Table~\ref{tab:cifar_ab}, we study the effect of using different normalizations on the generalization ability for CIFAR-10-C benchmark (Table~\ref{tab:mnist_ab} in Appendix reports the results on Digits). We observe that BN underperforms all the other normalization methods that have a consistent computational graph between training and testing. For this reason, our experiment with SN only includes IN and LN, but excludes BN. With the generic form that allows the model to adapt to single domain generalization easily, {\ours} outperforms BN, SN, and IN. The performance gain improves with the corruption level increasing. We also note that both AR and AS play a significant role in the performance gain, meaning learning both standardization and rescaling statistics in normalization is indeed helping models to learn to generalize. Further, AS achieves better performance than SN, which implies that learning only the combination weights for different standardization statistics are not as good as learning the statistics with neural networks. 
\begin{table}[ht] 
\vspace{-2mm}
\centering
\resizebox{0.95\columnwidth}{!}{
\begin{tabular}{@{}lccccccllllllll@{}}\toprule
Method & Level 1 & Level 2 & Level 3 & Level 4 & Level 5 & Avg.\\ \midrule
ADA+BN\cite{volpi2018generalizing} & 88.3 & 83.5 & 77.6 & 70.6 & 58.3 & 75.7\\
ADA+IN\cite{ulyanov2016instance} & 90.3 & 87.3 & 84.5 & 80.5 & 75.0 & 83.5 \\
ADA+SN\cite{luo2018differentiable} &	{\bf 91.5}&	88.4&	85.5&	81.2&	75.3&	84.4\\  \midrule
ADA+AR & 90.4 & 87.7 & 85.1 & 81.1 & 76.6 & 84.2 \\
ADA+AS & {91.4} & 88.9 & 86.3 & 82.8 & 77.3 & 85.4 \\
ADA+{ASR} (Ours) & {\bf 91.5} & {\bf 89.3} & {\bf 86.9} & {\bf 83.7} & {\bf 78.4} & {\bf 86.0}\\
\bottomrule
\end{tabular}}\vspace{-2mm}\caption{Single domain generalization accuracies on CIFAR-10-C with different normalizations. }\label{tab:cifar_ab}\vspace{-3mm}
\end{table}


{\bf Uncertainty Evaluation.} For CIFAR-10-C, we also evaluate the quality of {\it predictive probabilities} using 
{\it Brier score} (BS) \cite{brier1950verification},
defined as the squared distance between the predictive distribution and the one-hot target labels: 
$BS = \sum_{k=1}^K(1_{\{k=Y\}}-p_\theta(Y=k|X))^2/K$, where $K$ is the number of classes, and $p_\theta$ is the predictive probability. 
Table~\ref{tab:cifar_unc}
shows that {\ours} outperforms BN and IN in uncertainty prediction, meaning that {\ours} provides not only better predictions for classes but also better predictive distributions. The improvement is also increasing along with the rise of the domain discrepancy. 

\begin{table}[ht] 
\vspace{-2mm}
\centering
\resizebox{1\columnwidth}{!}{
\begin{tabular}{@{}lccccccllllllll@{}}\toprule
Method & Level 1 & Level 2 & Level 3 & Level 4 & Level 5 & Avg.\\ \midrule
ADA+BN\cite{volpi2018generalizing} & 0.019 & 0.028 & 0.035& 0.044 & 0.061& 0.037\\
ADA+IN\cite{ulyanov2016instance} & 0.015 & 0.020 & 0.025 & 0.032 & 0.041 & 0.027 \\ \hline
ADA+ASR (Ours) & {\bf 0.014} & {\bf 0.018} & {\bf 0.022} & {\bf 0.027} & {\bf 0.037} & {\bf 0.024}\\
{$\Delta$} to BN  & -0.005 & -0.010 & -0.013 & -0.017 & -0.024 & -0.013\\
\bottomrule
\end{tabular}}\vspace{-2mm}\caption{Uncertainty evaluations of different normalizations at different corruption levels using Brier score (the smaller the better) on CIFAR-10-C. }\label{tab:cifar_unc}\vspace{-3mm}
\end{table}

{\bf In-domain Performance and Adversarial Robustness.} In Table~\ref{tab:adv}, we report the in-domain performance of different normalizations by evaluating on the original CIFAR-10 testing set without corruptions. We note that BN indeed achieves better performance than IN and {\ours} on in-domain data. One explanation for that is the dependencies between training samples introduces the inductive biases that would help BN when samples come from the same domain. We also test the robustness towards adversarial attacks of each normalization, where we first apply several adversarial updates on each testing image and then make predictions on the adversarially corrupted image. The results in Table~\ref{tab:adv} show that ASR-Norm benefits from its adaptation capability and achieves better performance under adversarial attack. 

\begin{table}[ht] 
\vspace{-2mm}
\centering
\resizebox{0.8\columnwidth}{!}{
\begin{tabular}{@{}lccccclllllll@{}}\toprule
Domain & ADA+BN &ADA+IN &ADA+ASR (Ours) \\ \midrule
In-domain & {\bf 95.4} & 94.6 &  94.6\\
Adversarial & 32.0 & 46.9 & {\bf 52.2}\\
\bottomrule
\end{tabular}}\vspace{-2mm}\caption{Evaluation of different normalizations for in-domain data and adversarially corrupted images. }\label{tab:adv}\vspace{-3.0mm}
\end{table}

\begin{figure}[htp!]\vspace{-2mm} \centering
 \centering
 \includegraphics[width=0.6\linewidth]{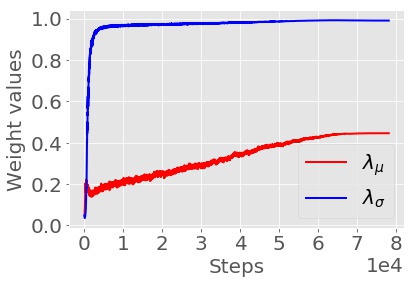}
 \vspace{-3mm}
\caption{\small Weights $\lambda_\muv$, $\lambda_\sigmav$ learn to increase along the training process for the first normalization layer in WRN on CIFAR-10-C.}
\label{fig:weight_adapt_only}\vspace{-5.5mm}
\end{figure}

{\bf Analysis of Residual Learning.} In our experiments, we find that the residual learning is necessary for stabilizing the training of {\ours}. Without this part, {\ours} would have numerical instabilities and yield NAN. Fig.~\ref{fig:weight_adapt_only} shows the evolution of the adaptive weights $\lambda_\muv$ and $\lambda_\sigmav$ in the residual terms along the training process of the CIFAR-10-C benchmark. The weights for learned statistics are initialized close to 0 and learn to increase gradually, meaning that the model 
favors the learned statistics more and more, which verifies that learned statistics are indeed helping the model. Interestingly, at the end of training, $\lambda_\sigmav$ adapts to almost $1$, i.e., the normalization learns to use only learned $\sigmav_\text{stan}$
(see more plots on PACS in Fig.~\ref{fig:rescale_adapt} in Appendix.) 

 \vspace{2mm}
{\bf Trade-off between Complexity and Performance.} The complexity and performance of {\ours} are impacted by the number of {\ours} layers used and the hidden dimension sizes, $C_\text{stan}$ and $C_\text{rescale}$. 
As in \cite{jia2019instance}, we fix $C_\text{rescale}$ to $16$. We conduct ablation study on PACS with the ResNet-18 backbone, shown in Table~\ref{tab:pacs_ablation}, to investigate the trade-off between complexity and performance by varying both the number of {\ours} layers applied and $C_\text{stan}$. First, we observe that a single layer of {\ours} improves over the baseline with only $0.1\%$ increase in parameters. As the number of {\ours} layers increases, the performance improves further at the expense of higher complexity. Second, applying {\ours} to all normalization layers with $C_\text{stan}=C/16$ could provide a significant performance boost over baseline with a small memory addition ($4.1\%$). Average accuracy increases as $C_\text{stan}$ increases with a sub-linear speed while parameters increase linearly with $C_\text{stan}$.

\begin{table}[ht] 
\vspace{-2mm}
\centering
\resizebox{1\columnwidth}{!}{
\begin{tabular}{@{}lccccclllllll@{}}\toprule
Norm layers ($C_\text{stan}$, $C_\text{rescale}$) & Avg. & \# Params & Params $\uparrow$ \\ \midrule
BN  & 61.8 & 11.18M & 0 \\ 
ASR to the first norm layer ($C/2,C/16$) & 62.3&11.19M & 0.1\% \\
ASR to selected 4 norm layers ($C/2,C/16$)  & 64.4 & 11.80M & 5.5\% \\
ASR to all norm layers ($C/16,C/16$)&  66.5 &11.64M & 4.1\%\\
ASR to all norm layers ($C/8,C/16$)& 67.5 &11.83M & 5.8\%\\
ASR to all norm layers ($C/2,C/16$) &{68.1} & 13.01M & 16.4\%\\
\bottomrule
\end{tabular}}\vspace{-2mm}\caption{Trade-off between model complexity and accuracies on PACS. One domain is used for training. Average accuracies, 
parameter sizes, and the increasing proportion of parameter sizes with respect to the baseline are reported. 
}\label{tab:pacs_ablation}\vspace{-5mm}
\end{table}



\section{Conclusion}
We propose {\ours}, a novel adaptive and general form of normalization which learns both the standardization and rescaling statistics with auto-encoder structured neural networks. {\ours} is a generic algorithm that complements a variety of adversarial domain augmentation (ADA) approaches on single domain generalization by making the statistics be adaptive to the data of different domains, and hence improves the model generalization capability across domains. On three standard 
benchmarks, we show that ASR-Norm brings consistent improvement to the state-of-the-art ADA approaches. The performance gain is agnostic to the choice of ADA methods and becomes more significant as the domain discrepancy increases.

{\footnotesize {\bf Acknowledgements. } X. Fan and M. Zhou acknowledge the support of NSF IIS-1812699 and Texas Advanced Computing Center.}
\clearpage
{\small
\setlength{\bibsep}{0pt}
\bibliographystyle{abbrvnat}
\bibliography{egbib}

\begin{thebibliography}{61}
\providecommand{\natexlab}[1]{#1}
\providecommand{\url}[1]{\texttt{#1}}
\expandafter\ifx\csname urlstyle\endcsname\relax
  \providecommand{\doi}[1]{doi: #1}\else
  \providecommand{\doi}{doi: \begingroup \urlstyle{rm}\Url}\fi

\bibitem[Ba et~al.(2016)Ba, Kiros, and Hinton]{ba2016layer}
J.~L. Ba, J.~R. Kiros, and G.~E. Hinton.
\newblock Layer normalization.
\newblock \emph{arXiv preprint arXiv:1607.06450}, 2016.

\bibitem[Balaji et~al.(2018)Balaji, Sankaranarayanan, and
  Chellappa]{balaji2018metareg}
Y.~Balaji, S.~Sankaranarayanan, and R.~Chellappa.
\newblock Metareg: Towards domain generalization using meta-regularization.
\newblock In \emph{Advances in Neural Information Processing Systems}, pages
  998--1008, 2018.

\bibitem[Boyd et~al.(2004)Boyd, Boyd, and Vandenberghe]{boyd2004convex}
S.~Boyd, S.~P. Boyd, and L.~Vandenberghe.
\newblock \emph{Convex optimization}.
\newblock Cambridge university press, 2004.

\bibitem[Brier(1950)]{brier1950verification}
G.~W. Brier.
\newblock Verification of forecasts expressed in terms of probability.
\newblock \emph{Monthly weather review}, 78\penalty0 (1):\penalty0 1--3, 1950.

\bibitem[Carlucci et~al.(2019)Carlucci, D'Innocente, Bucci, Caputo, and
  Tommasi]{carlucci2019domain}
F.~M. Carlucci, A.~D'Innocente, S.~Bucci, B.~Caputo, and T.~Tommasi.
\newblock Domain generalization by solving jigsaw puzzles.
\newblock In \emph{Proceedings of the IEEE Conference on Computer Vision and
  Pattern Recognition}, pages 2229--2238, 2019.

\bibitem[Chattopadhyay et~al.(2020)Chattopadhyay, Balaji, and
  Hoffman]{chattopadhyay2020learning}
P.~Chattopadhyay, Y.~Balaji, and J.~Hoffman.
\newblock Learning to balance specificity and invariance for in and out of
  domain generalization.
\newblock \emph{ECCV}, 2020.

\bibitem[Chen et~al.(2020)Chen, Kornblith, Swersky, Norouzi, and
  Hinton]{chen2020big}
T.~Chen, S.~Kornblith, K.~Swersky, M.~Norouzi, and G.~Hinton.
\newblock Big self-supervised models are strong semi-supervised learners.
\newblock \emph{arXiv e-prints}, pages arXiv--2006, 2020.

\bibitem[DeVries and Taylor(2017)]{devries2017improved}
T.~DeVries and G.~W. Taylor.
\newblock Improved regularization of convolutional neural networks with cutout.
\newblock \emph{arXiv preprint arXiv:1708.04552}, 2017.

\bibitem[Fan et~al.(2020)Fan, Zhang, Chen, and Zhou]{fan2020bayesian}
X.~Fan, S.~Zhang, B.~Chen, and M.~Zhou.
\newblock Bayesian attention modules.
\newblock In \emph{Advances in Neural Information Processing Systems},
  volume~33, pages 16362--16376, 2020.

\bibitem[Fan et~al.(2021)Fan, Zhang, Tanwisuth, Qian, and
  Zhou]{fan2021contextual}
X.~Fan, S.~Zhang, K.~Tanwisuth, X.~Qian, and M.~Zhou.
\newblock Contextual dropout: An efficient sample-dependent dropout module.
\newblock In \emph{International Conference on Learning Representations}, 2021.

\bibitem[Ganin and Lempitsky(2015)]{ganin2015unsupervised}
Y.~Ganin and V.~Lempitsky.
\newblock Unsupervised domain adaptation by backpropagation.
\newblock In \emph{International conference on machine learning}, pages
  1180--1189. PMLR, 2015.

\bibitem[Ghiasi et~al.(2018)Ghiasi, Lin, and Le]{ghiasi2018dropblock}
G.~Ghiasi, T.-Y. Lin, and Q.~V. Le.
\newblock Dropblock: A regularization method for convolutional networks.
\newblock In \emph{Advances in Neural Information Processing Systems}, pages
  10727--10737, 2018.

\bibitem[Goodfellow et~al.(2014)Goodfellow, Shlens, and
  Szegedy]{goodfellow2014explaining}
I.~J. Goodfellow, J.~Shlens, and C.~Szegedy.
\newblock Explaining and harnessing adversarial examples.
\newblock \emph{arXiv preprint arXiv:1412.6572}, 2014.

\bibitem[Hendrycks and Dietterich(2018)]{hendrycks2018benchmarking}
D.~Hendrycks and T.~Dietterich.
\newblock Benchmarking neural network robustness to common corruptions and
  perturbations.
\newblock In \emph{International Conference on Learning Representations}, 2018.

\bibitem[Hendrycks et~al.(2019{\natexlab{a}})Hendrycks, Mazeika, Kadavath, and
  Song]{hendrycks2019using}
D.~Hendrycks, M.~Mazeika, S.~Kadavath, and D.~Song.
\newblock Using self-supervised learning can improve model robustness and
  uncertainty.
\newblock In \emph{Advances in Neural Information Processing Systems}, pages
  15663--15674, 2019{\natexlab{a}}.

\bibitem[Hendrycks et~al.(2019{\natexlab{b}})Hendrycks, Mu, Cubuk, Zoph,
  Gilmer, and Lakshminarayanan]{hendrycks2019augmix}
D.~Hendrycks, N.~Mu, E.~D. Cubuk, B.~Zoph, J.~Gilmer, and B.~Lakshminarayanan.
\newblock Augmix: A simple data processing method to improve robustness and
  uncertainty.
\newblock In \emph{International Conference on Learning Representations},
  2019{\natexlab{b}}.

\bibitem[Hendrycks et~al.(2020{\natexlab{a}})Hendrycks, Basart, Mu, Kadavath,
  Wang, Dorundo, Desai, Zhu, Parajuli, Guo, et~al.]{hendrycks2020many}
D.~Hendrycks, S.~Basart, N.~Mu, S.~Kadavath, F.~Wang, E.~Dorundo, R.~Desai,
  T.~Zhu, S.~Parajuli, M.~Guo, et~al.
\newblock The many faces of robustness: A critical analysis of
  out-of-distribution generalization.
\newblock \emph{arXiv preprint arXiv:2006.16241}, 2020{\natexlab{a}}.

\bibitem[Hendrycks et~al.(2020{\natexlab{b}})Hendrycks, Liu, Wallace, Dziedzic,
  Krishnan, and Song]{hendrycks2020pretrained}
D.~Hendrycks, X.~Liu, E.~Wallace, A.~Dziedzic, R.~Krishnan, and D.~Song.
\newblock Pretrained transformers improve out-of-distribution robustness.
\newblock \emph{arXiv preprint arXiv:2004.06100}, 2020{\natexlab{b}}.

\bibitem[Hu et~al.(2018)Hu, Shen, and Sun]{hu2018squeeze}
J.~Hu, L.~Shen, and G.~Sun.
\newblock Squeeze-and-excitation networks.
\newblock In \emph{Proceedings of the IEEE conference on computer vision and
  pattern recognition}, pages 7132--7141, 2018.

\bibitem[Huang et~al.(2020)Huang, Wang, Xing, and Huang]{huang2020self}
Z.~Huang, H.~Wang, E.~P. Xing, and D.~Huang.
\newblock Self-challenging improves cross-domain generalization.
\newblock \emph{ECCV}, 2020.

\bibitem[Hull(1994)]{hull1994database}
J.~J. Hull.
\newblock A database for handwritten text recognition research.
\newblock \emph{IEEE Transactions on pattern analysis and machine
  intelligence}, 16\penalty0 (5):\penalty0 550--554, 1994.

\bibitem[Ioffe and Szegedy(2015)]{ioffe2015batch}
S.~Ioffe and C.~Szegedy.
\newblock Batch normalization: Accelerating deep network training by reducing
  internal covariate shift.
\newblock In \emph{International Conference on Machine Learning}, pages
  448--456, 2015.

\bibitem[Jia et~al.(2019)Jia, Chen, and Chen]{jia2019instance}
S.~Jia, D.-J. Chen, and H.-T. Chen.
\newblock Instance-level meta normalization.
\newblock In \emph{Proceedings of the IEEE conference on computer vision and
  pattern recognition}, pages 4865--4873, 2019.

\bibitem[Kingma and Ba(2014)]{kingma2014adam}
D.~P. Kingma and J.~Ba.
\newblock Adam: A method for stochastic optimization.
\newblock \emph{arXiv preprint arXiv:1412.6980}, 2014.

\bibitem[Krizhevsky et~al.(2017)Krizhevsky, Sutskever, and
  Hinton]{krizhevsky2017imagenet}
A.~Krizhevsky, I.~Sutskever, and G.~E. Hinton.
\newblock Imagenet classification with deep convolutional neural networks.
\newblock \emph{Communications of the ACM}, 60\penalty0 (6):\penalty0 84--90,
  2017.

\bibitem[Krizhevsky et~al.(2009)]{krizhevsky2009learning}
A.~Krizhevsky et~al.
\newblock Learning multiple layers of features from tiny images.
\newblock Technical report, Citeseer, 2009.

\bibitem[LeCun et~al.(1989)LeCun, Boser, Denker, Henderson, Howard, Hubbard,
  and Jackel]{lecun1989backpropagation}
Y.~LeCun, B.~Boser, J.~S. Denker, D.~Henderson, R.~E. Howard, W.~Hubbard, and
  L.~D. Jackel.
\newblock Backpropagation applied to handwritten zip code recognition.
\newblock \emph{Neural computation}, 1\penalty0 (4):\penalty0 541--551, 1989.

\bibitem[LeCun et~al.(2015)LeCun, Bengio, and Hinton]{lecun2015deep}
Y.~LeCun, Y.~Bengio, and G.~Hinton.
\newblock Deep learning.
\newblock \emph{nature}, 521\penalty0 (7553):\penalty0 436--444, 2015.

\bibitem[Li et~al.(2020)Li, Wu, Lim, Belongie, and Weinberger]{li2020feature}
B.~Li, F.~Wu, S.-N. Lim, S.~Belongie, and K.~Q. Weinberger.
\newblock On feature normalization and data augmentation.
\newblock \emph{arXiv preprint arXiv:2002.11102}, 2020.

\bibitem[Li et~al.(2017)Li, Yang, Song, and Hospedales]{li2017deeper}
D.~Li, Y.~Yang, Y.-Z. Song, and T.~M. Hospedales.
\newblock Deeper, broader and artier domain generalization.
\newblock In \emph{Proceedings of the IEEE international conference on computer
  vision}, pages 5542--5550, 2017.

\bibitem[Loshchilov and Hutter(2016)]{loshchilov2016sgdr}
I.~Loshchilov and F.~Hutter.
\newblock Sgdr: Stochastic gradient descent with warm restarts.
\newblock \emph{arXiv preprint arXiv:1608.03983}, 2016.

\bibitem[Luo et~al.(2018)Luo, Ren, Peng, Zhang, and Li]{luo2018differentiable}
P.~Luo, J.~Ren, Z.~Peng, R.~Zhang, and J.~Li.
\newblock Differentiable learning-to-normalize via switchable normalization.
\newblock In \emph{International Conference on Learning Representations}, 2018.

\bibitem[Maaten and Hinton(2008)]{maaten2008visualizing}
L.~v.~d. Maaten and G.~Hinton.
\newblock Visualizing data using t-sne.
\newblock \emph{Journal of machine learning research}, 9\penalty0
  (Nov):\penalty0 2579--2605, 2008.

\bibitem[Madry et~al.(2017)Madry, Makelov, Schmidt, Tsipras, and
  Vladu]{madry2017towards}
A.~Madry, A.~Makelov, L.~Schmidt, D.~Tsipras, and A.~Vladu.
\newblock Towards deep learning models resistant to adversarial attacks.
\newblock \emph{arXiv preprint arXiv:1706.06083}, 2017.

\bibitem[Madry et~al.(2018)Madry, Makelov, Schmidt, Tsipras, and
  Vladu]{madry2018towards}
A.~Madry, A.~Makelov, L.~Schmidt, D.~Tsipras, and A.~Vladu.
\newblock Towards deep learning models resistant to adversarial attacks.
\newblock In \emph{International Conference on Learning Representations}, 2018.

\bibitem[Motiian et~al.(2017)Motiian, Piccirilli, Adjeroh, and
  Doretto]{motiian2017unified}
S.~Motiian, M.~Piccirilli, D.~A. Adjeroh, and G.~Doretto.
\newblock Unified deep supervised domain adaptation and generalization.
\newblock In \emph{Proceedings of the IEEE International Conference on Computer
  Vision}, pages 5715--5725, 2017.

\bibitem[Nado et~al.(2020)Nado, Padhy, Sculley, D'Amour, Lakshminarayanan, and
  Snoek]{nado2020evaluating}
Z.~Nado, S.~Padhy, D.~Sculley, A.~D'Amour, B.~Lakshminarayanan, and J.~Snoek.
\newblock Evaluating prediction-time batch normalization for robustness under
  covariate shift.
\newblock \emph{arXiv preprint arXiv:2006.10963}, 2020.

\bibitem[Nair and Hinton(2010)]{nair2010rectified}
V.~Nair and G.~E. Hinton.
\newblock Rectified linear units improve restricted boltzmann machines.
\newblock In \emph{ICML}, 2010.

\bibitem[Nam and Kim(2018)]{nam2018batch}
H.~Nam and H.-E. Kim.
\newblock Batch-instance normalization for adaptively style-invariant neural
  networks.
\newblock In \emph{Advances in Neural Information Processing Systems}, pages
  2558--2567, 2018.

\bibitem[Netzer et~al.(2011)Netzer, Wang, Coates, Bissacco, Wu, and
  Ng]{netzer2011reading}
Y.~Netzer, T.~Wang, A.~Coates, A.~Bissacco, B.~Wu, and A.~Y. Ng.
\newblock Reading digits in natural images with unsupervised feature learning.
\newblock 2011.

\bibitem[Park et~al.(2017)Park, Park, Shin, and Moon]{park2017adversarial}
S.~Park, J.-K. Park, S.-J. Shin, and I.-C. Moon.
\newblock Adversarial dropout for supervised and semi-supervised learning.
\newblock \emph{arXiv preprint arXiv:1707.03631}, 2017.

\bibitem[Qiao et~al.(2020)Qiao, Zhao, and Peng]{qiao2020learning}
F.~Qiao, L.~Zhao, and X.~Peng.
\newblock Learning to learn single domain generalization.
\newblock In \emph{Proceedings of the IEEE/CVF Conference on Computer Vision
  and Pattern Recognition}, pages 12556--12565, 2020.

\bibitem[Saenko et~al.(2010)Saenko, Kulis, Fritz, and
  Darrell]{saenko2010adapting}
K.~Saenko, B.~Kulis, M.~Fritz, and T.~Darrell.
\newblock Adapting visual category models to new domains.
\newblock In \emph{European conference on computer vision}, pages 213--226.
  Springer, 2010.

\bibitem[Salimans et~al.(2016)Salimans, Goodfellow, Zaremba, Cheung, Radford,
  and Chen]{salimans2016improved}
T.~Salimans, I.~Goodfellow, W.~Zaremba, V.~Cheung, A.~Radford, and X.~Chen.
\newblock Improved techniques for training gans.
\newblock In \emph{Advances in neural information processing systems}, pages
  2234--2242, 2016.

\bibitem[Seo et~al.(2019)Seo, Suh, Kim, Han, and Han]{seo2019learning}
S.~Seo, Y.~Suh, D.~Kim, J.~Han, and B.~Han.
\newblock Learning to optimize domain specific normalization for domain
  generalization.
\newblock \emph{ECCV}, 2019.

\bibitem[Shao et~al.(2019)Shao, Meng, Li, Zhang, Li, Wang, and
  Luo]{shao2019ssn}
W.~Shao, T.~Meng, J.~Li, R.~Zhang, Y.~Li, X.~Wang, and P.~Luo.
\newblock Ssn: Learning sparse switchable normalization via sparsestmax.
\newblock In \emph{Proceedings of the IEEE Conference on Computer Vision and
  Pattern Recognition}, pages 443--451, 2019.

\bibitem[Sinha et~al.(2018)Sinha, Namkoong, and Duchi]{sinha2018certifying}
A.~Sinha, H.~Namkoong, and J.~Duchi.
\newblock Certifying some distributional robustness with principled adversarial
  training.
\newblock In \emph{International Conference on Learning Representations}, 2018.

\bibitem[Sun et~al.(2020)Sun, Wang, Liu, Miller, Efros, and Hardt]{sun2020test}
Y.~Sun, X.~Wang, Z.~Liu, J.~Miller, A.~Efros, and M.~Hardt.
\newblock Test-time training with self-supervision for generalization under
  distribution shifts.
\newblock In \emph{International Conference on Machine Learning}, pages
  9229--9248. PMLR, 2020.

\bibitem[Tzeng et~al.(2017)Tzeng, Hoffman, Saenko, and
  Darrell]{tzeng2017adversarial}
E.~Tzeng, J.~Hoffman, K.~Saenko, and T.~Darrell.
\newblock Adversarial discriminative domain adaptation.
\newblock In \emph{Proceedings of the IEEE conference on computer vision and
  pattern recognition}, pages 7167--7176, 2017.

\bibitem[Ulyanov et~al.(2016)Ulyanov, Vedaldi, and
  Lempitsky]{ulyanov2016instance}
D.~Ulyanov, A.~Vedaldi, and V.~Lempitsky.
\newblock Instance normalization: The missing ingredient for fast stylization.
\newblock \emph{arXiv preprint arXiv:1607.08022}, 2016.

\bibitem[Vapnik(2013)]{vapnik2013nature}
V.~Vapnik.
\newblock \emph{The nature of statistical learning theory}.
\newblock Springer science \& business media, 2013.

\bibitem[Volpi and Murino(2019)]{volpi2019addressing}
R.~Volpi and V.~Murino.
\newblock Addressing model vulnerability to distributional shifts over image
  transformation sets.
\newblock In \emph{Proceedings of the IEEE International Conference on Computer
  Vision}, pages 7980--7989, 2019.

\bibitem[Volpi et~al.(2018)Volpi, Namkoong, Sener, Duchi, Murino, and
  Savarese]{volpi2018generalizing}
R.~Volpi, H.~Namkoong, O.~Sener, J.~C. Duchi, V.~Murino, and S.~Savarese.
\newblock Generalizing to unseen domains via adversarial data augmentation.
\newblock In \emph{Advances in neural information processing systems}, pages
  5334--5344, 2018.

\bibitem[Wang et~al.(2019)Wang, Jin, Long, Wang, and
  Jordan]{wang2019transferable}
X.~Wang, Y.~Jin, M.~Long, J.~Wang, and M.~I. Jordan.
\newblock Transferable normalization: Towards improving transferability of deep
  neural networks.
\newblock In \emph{Advances in Neural Information Processing Systems}, pages
  1953--1963, 2019.

\bibitem[Wu and He(2018)]{wu2018group}
Y.~Wu and K.~He.
\newblock Group normalization.
\newblock In \emph{Proceedings of the European conference on computer vision
  (ECCV)}, pages 3--19, 2018.

\bibitem[Xu et~al.(2019{\natexlab{a}})Xu, Sun, Zhang, Zhao, and
  Lin]{xu2019understanding}
J.~Xu, X.~Sun, Z.~Zhang, G.~Zhao, and J.~Lin.
\newblock Understanding and improving layer normalization.
\newblock In \emph{Advances in Neural Information Processing Systems}, pages
  4381--4391, 2019{\natexlab{a}}.

\bibitem[Xu et~al.(2019{\natexlab{b}})Xu, Zhou, Venkatesan, Swaminathan, and
  Majumder]{xu2019d}
X.~Xu, X.~Zhou, R.~Venkatesan, G.~Swaminathan, and O.~Majumder.
\newblock d-sne: Domain adaptation using stochastic neighborhood embedding.
\newblock In \emph{Proceedings of the IEEE conference on computer vision and
  pattern recognition}, pages 2497--2506, 2019{\natexlab{b}}.

\bibitem[Zagoruyko and Komodakis(2016)]{zagoruyko2016wide}
S.~Zagoruyko and N.~Komodakis.
\newblock Wide residual networks.
\newblock \emph{arXiv preprint arXiv:1605.07146}, 2016.

\bibitem[Zhao et~al.(2018)Zhao, Zhang, Wu, Moura, Costeira, and
  Gordon]{zhao2018adversarial}
H.~Zhao, S.~Zhang, G.~Wu, J.~M. Moura, J.~P. Costeira, and G.~J. Gordon.
\newblock Adversarial multiple source domain adaptation.
\newblock In \emph{Advances in neural information processing systems}, pages
  8559--8570, 2018.

\bibitem[Zhao et~al.(2020)Zhao, Liu, Peng, and Metaxas]{zhao2020maximum}
L.~Zhao, T.~Liu, X.~Peng, and D.~Metaxas.
\newblock Maximum-entropy adversarial data augmentation for improved
  generalization and robustness.
\newblock \emph{arXiv preprint arXiv:2010.08001}, 2020.

\bibitem[Zhou et~al.(2020)Zhou, Yang, Hospedales, and Xiang]{zhou2020learning}
K.~Zhou, Y.~Yang, T.~Hospedales, and T.~Xiang.
\newblock Learning to generate novel domains for domain generalization.
\newblock \emph{arXiv preprint arXiv:2007.03304}, 2020.

\end{thebibliography}
}

\clearpage
\appendix

\begin{center}
  \Large{\bf Appendices for ``Adversarially Adaptive Normalization for Single Domain Generalization'' }  
\end{center}

\section{Additional Experimental Results}
\label{sec:app_results}

{\bf On the Effect of Normalization.} In Table~\ref{tab:mnist_ab}, we study the impact on the generalization ability by using different normalization techniques with RSDA \cite{volpi2018generalizing} for domain augmentation on the Digits benchmark. 
It reports average accuracies, standard deviations and $p$-values 
for each domain in the Digits benchmark. 
We observe that adding either batch normalization (BN) or BN-test to the ConvNet architecture makes the performance worse than the baseline without any normalization layer. Instance normalization shows small improvement over the baseline but still underperforms {\ours}. {\ours} outperforms all methods on average and achieves significant improvement for challenging domains, including SVHN and SYN. On the easier domains like MNIST-M and USPS, ASR performs on a par with the baseline (RSDA).

\begin{table}[ht] 
\vspace{-2mm}
\centering
\resizebox{0.95\columnwidth}{!}{
\begin{tabular}{@{}lccccclllllll@{}}\toprule
Method & SVHN & MNIST-M & SYN & USPS & Avg.\\ \midrule
RSDA+BN\cite{ioffe2015batch} & 39.4$\pm$5.2 & 76.6$\pm$2.2 & 60.5$\pm$1.5 & 84.2$\pm$2.1 & 65.2 \\ 
RSDA+BN-Test\cite{nado2020evaluating} & 45.7$\pm$2.8 & 80.3$\pm$1.2 & 59.7$\pm$1.4 & 81.8$\pm$1.1 & 66.9 \\ 
RSDA+IN\cite{ulyanov2016instance} & 47.1$\pm$3.4 & 80.6$\pm$0.9 & 61.9$\pm$1.5 & 85.4$\pm$1.4 & 68.8 \\
RSDA+SN\cite{luo2018differentiable} & 37.7$\pm$3.8 & 77.1$\pm$1.4 & 60.5$\pm$1.8 & 86.1$\pm$1.7 & 65.4 \\
RSDA & 47.4$\pm$4.8 & {
81.5}$\pm$1.6 & 62.0$\pm$1.2 & 83.1$\pm$1.2 &68.5\\ 
RSDA+AR & 47.8 $\pm$3.2 & 80.0$\pm$1.0 & 64.0$\pm$0.9 & {\bf 86.7$\pm$1.5} &69.6\\
RSDA+AS & 49.4$\pm$2.3 &	81.4$\pm$0.7&	63.5$\pm$1.2&	81.4$\pm$1.1&	69.3\\
RSDA+ASR (Ours) & {\bf 52.8$\pm$3.8} & {80.8}$\pm$0.6 & {\bf 64.5$\pm$1.1} & 82.4$\pm$1.4& {\bf 70.1} \\ \midrule
$p$-value: Ours vs. RSDA & {\bf 0.036} & 0.193 & {\bf 0.003} & 0.197  & - \\
$p$-value: Ours vs. AS &{\bf 0.050}&	0.088&	0.115	&0.108 & - \\
$p$-value: Ours vs. AR & {\bf 0.020} & 0.080 & 0.214 & $\mathbf{<1e\minus 3}$ & - \\
\bottomrule
\end{tabular}}\vspace{3mm}\caption{Single domain generalization accuracies with different normalization on Digits. MNIST is used as the training set, and the results on different testing domains are reported in different columns.}\label{tab:mnist_ab}
\end{table}



{\bf Statistical significance of results on CIFAR-10-C.} In Table~\ref{tab:cifar_pvalue} reports the standard deviations and $p$-values for the one-sided two-sample $t$-test on the accuracies for CIFAR-10-C in addition to Table~\ref{tab:cifar_ab}. The results show consistently statistical significance of ASR-norm's improvements over M-ADA, SN, AR, and AS in different corruption levels.

\begin{table}[t] 
\vspace{0mm}
\centering
\resizebox{0.99\columnwidth}{!}{
\begin{tabular}{@{}lccccccllllllll@{}}\toprule
Method & Level 1 & Level 2 & Level 3 & Level 4 & Level 5 & Avg.\\ \midrule
ERM+BN & 87.8$\pm$0.1 & 81.5$\pm$0.2 & 75.5$\pm$0.4 & 68.2$\pm$0.6 & 56.1$\pm$0.8 & 73.8 \\
ERM+ASR (ASR alone) & 89.4$\pm$0.2 & 86.1$\pm$0.2 & 82.9$\pm$0.3 & 78.6$\pm$0.6 & 72.9$\pm$1.0&82.0  \\
M-ADA & 90.5$\pm$0.3 & 86.8$\pm$0.4 & 82.5$\pm$0.6 & 76.4$\pm$0.9 & 65.6$\pm$1.2  &80.4\\
ADA+SN &	{\bf 91.5$\pm$0.2}&	88.4$\pm$0.6&	85.5$\pm$0.5&	81.2$\pm$0
7&	75.3$\pm$0.8 &84.4\\  
ADA+AR & 90.4$\pm$0.1 & 87.7$\pm$0.3 & 85.1$\pm$0.6 & 81.1$\pm$0.7 & 76.6$\pm$1.0 & 84.2\\
ADA+AS & {91.4}$\pm$0.1 & 88.9$\pm$0.2 & 86.3$\pm$0.4 & 82.8$\pm$0.5 & 77.3$\pm$0.7  & 85.4\\
ADA+ASR (Ours) & {\bf 91.5$\pm$0.2} & {\bf 89.3$\pm$0.6} & {\bf 86.9$\pm$0.5} & {\bf 83.7$\pm$0.7} & {\bf 78.4$\pm$0.8} &{\bf 86.0}\\
\midrule
$p$-value: Ours vs. \footnotesize{ERM/ERM+ASR/M-ADA} &  $\mathbf{<1e\minus 3}$ & $\mathbf{<1e\minus 3}$ & $\mathbf{<1e\minus 3}$ & $\mathbf{<1e\minus 3}$ & $\mathbf{<1e\minus 3}$ & -\\
$p$-value: Ours vs. ADA+SN &  0.5 & $\mathbf{0.025}$ & $\mathbf{0.006}$ & $\mathbf{0
.001}$ & $\mathbf{<1e\minus 3}$&-\\
$p$-value: Ours vs. ADA+AR &  $\mathbf{0.001}$ & $\mathbf{0.003}$ & $\mathbf{0.005}$ & $\mathbf{0.003}$ & $\mathbf{0.050}$ & -\\
$p$-value: Ours vs. ADA+AS &  ${0.199}$ & ${0.121}$ & $\mathbf{0.049}$ & $\mathbf{0.035}$ & $\mathbf{0.036}$ & -\\
\bottomrule
\end{tabular}}\vspace{-1mm}\caption{Single domain generalization accuracies and $p$-values on CIFAR-10-C with different corruption levels. Significant results are highlighted ($p$-value $\leq 0.05$).}\label{tab:cifar_pvalue}\vspace{-3mm}
\end{table}

\begin{figure*}[th!]\vspace{-6mm}
\centering
\includegraphics[height=7.2cm]{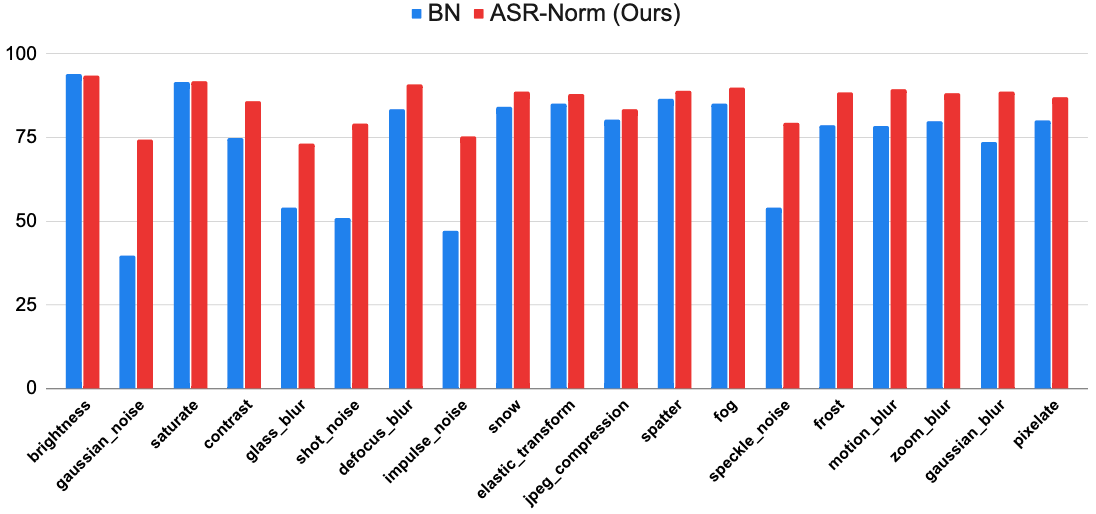}\vspace{-4mm}
\caption{\small Single domain generalization results on CIFAR-10-C for each corruption type. }
\label{fig:cifar_type} \vspace{-1.5mm}
\end{figure*} 

\begin{figure}[htp!]
\centering
\begin{subfigure}[t]{.247\textwidth}
 \centering
 \includegraphics[width=1\linewidth]{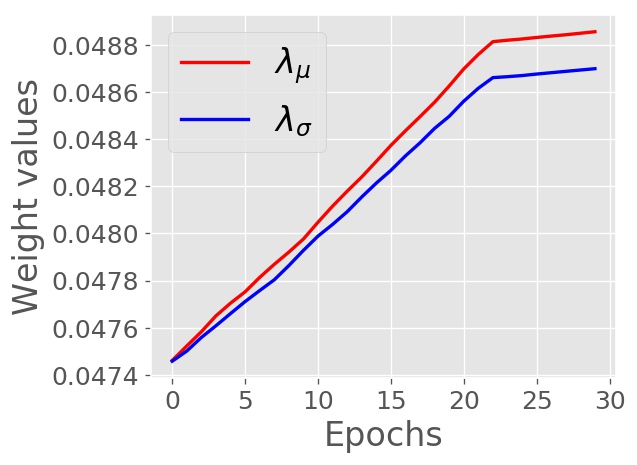}
 \subcaption[a]{Weights $\lambda_\muv, \lambda_\sigmav$.}\label{fig:weight_stan}
\end{subfigure}
\begin{subfigure}[t]{.247\textwidth}
 \centering
 \includegraphics[width=1\linewidth]{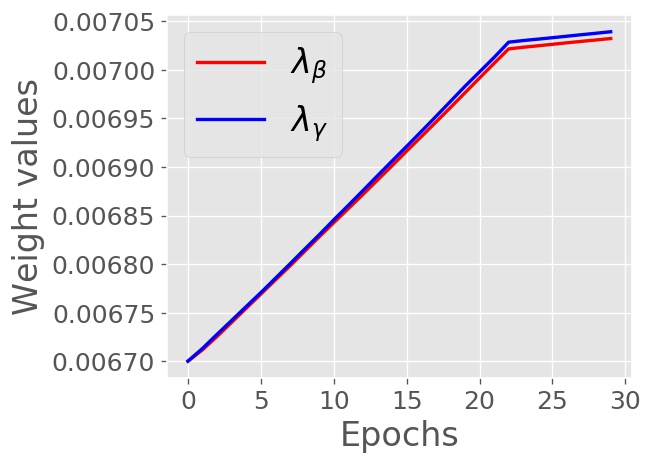}
  \subcaption[b]{Weights $\lambda_\beta, \lambda_\gamma$.}\label{fig:weight_rescale}
 \vspace{1mm}
\end{subfigure}
\caption{\small Weights learn to increase the contribution from learned statistics along the training process on the PACS benchmark.}
\label{fig:rescale_adapt} \vspace{-1.5mm}
\end{figure}

{\bf Analysis of Residual Learning.}  Fig.~\ref{fig:weight_stan} shows the evolution of the adaptive weights $\lambda_\muv$ and $\lambda_\sigmav$ in the residual terms of standardization statistics along the training process of the PACS benchmark. The weights for learned statistics are initialized close to 0 and learn to increase gradually, meaning that the model 
favors the learned statistics increasingly along the training process. That verifies the learned statistics are indeed favored the model for domain generalization. We note that the increasing speed of the residual weights for PACS is not as fast as that for CIFAR-10-C. The reason for that could be we used a pretrained model for PACS, which already learned some useful statistics. Fig.~\ref{fig:weight_rescale} shows the evolution of the adaptive weights $\lambda_\beta$ and $\lambda_\gamma$ in the residual terms of rescaling statistics, where we have the similar observations.

\begin{figure}[htp!]\vspace{-2mm} \centering
\begin{subfigure}[t]{.247\textwidth}
 \centering
 \includegraphics[width=1\linewidth]{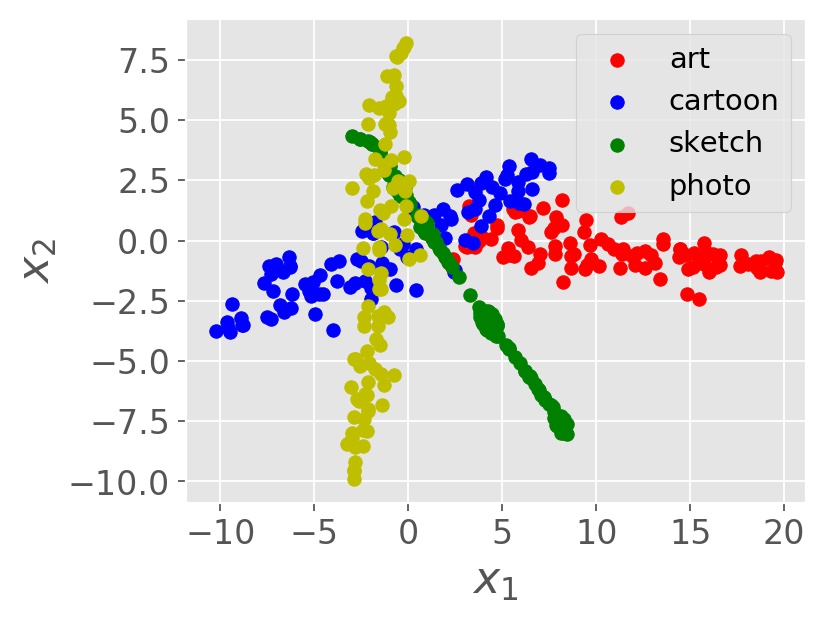}
\subcaption[a]{Visualization of $\sigmav_\text{stan}$.}\label{fig:vis_stan_sigma}
\end{subfigure}
\begin{subfigure}[t]{.247\textwidth}
 \centering
 \includegraphics[width=1\linewidth]{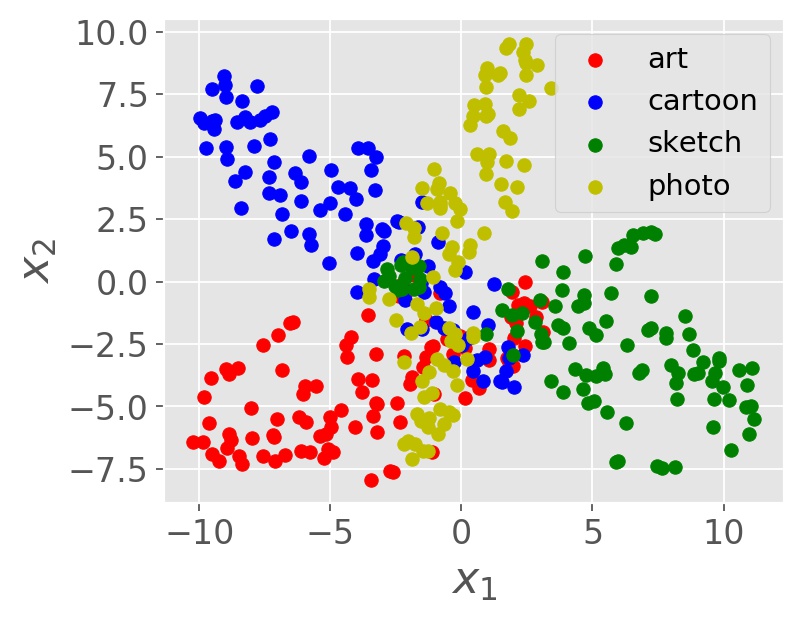}
 \subcaption[b]{Visualization of $\muv_\text{stan}$.}\label{fig:vis_stan_mu}
\end{subfigure}
 \vspace{-2mm}
\caption{\small Visualization of learned standardization statistics for different domains on PACS benchmark. }
\label{fig:tsne}\vspace{-3mm}
\end{figure}

{\bf Visualization of Learned Statistics. }Figure~\ref{fig:tsne} visualizes the learned standardization statistics $\muv_\text{stan}$ and $\sigmav_\text{stan}$ using t-SNE \cite{maaten2008visualizing} for different domains in PACS. We notice that the learned statistics show clustering structures for each domain, meaning that {\ours} learns different patterns of standardization statistics for each domain. This finding resembles previous papers on using domain-specific statistics for multi-domain data \cite{seo2019learning}. However, our method learns the soft clustered embeddings in an automatic way without the hard domain label on each sample. 

{\bf CIFAR-10-C Results for Different Corruption Types.} CIFAR-10-C contains $19$ corruption types including, brightness, gaussian noise, saturate, contrast, glass blur, shot noise, defocus blur, impulse noise, snow, elastic transform, jpeg compression, spatter, fog, speckle noise, frost, motion blur, zoom blur, gaussian blur, and pixelate. 
These $19$ corruption types can be categoried into $4$ categories including, noise, blur, weather, and digital categories \cite{hendrycks2018benchmarking}. Figure~\ref{fig:cifar_type} shows the average accuracies for each corruption type across five intensity levels. We observe that {\ours} makes consistent improvements over BN in most corruption types, except for brightness.

\section{Detailed Formulation of Adversarial Domain Augmentation}
\label{sec:app_ada}
Adversarial domain augmentation ({ADA}) \cite{volpi2018generalizing} approximately optimizes the robust objective $\mathcal{L}_{RL}$ in Eq \ref{eq:obj_rl} by expanding the training set with synthesized adversarial examples along the training process. 
Specifically, we define the distance $D$ between two distributions $P$ and $Q$ by the Wasserstein distance as \cite{volpi2018generalizing}:
\begin{equation}
    D_\theta(P, Q):= \inf _{M\in \Pi (P, Q)} \E _M [c_\theta ((X,Y), (X', Y'))],
\end{equation}
where $c_\theta$ is a learned distance measure over the space $\mathcal{X}\times \mathcal{Y}$. In ADA, $c_\theta$ is measured with the semantic features learned by the neural networks: 
\begin{equation}
    c_\theta((x,y), (x', y')):=c((F_\theta(x), y), (F_\theta(x'), y')),
\end{equation}
where $F_\theta$ is a feature extractor outputting intermediate activations in the neural networks, and 
\begin{equation}
    c((z,y), (z', y')) :=\frac{1}{2} \left\lVert z-z' \right\rVert_2^2+\infty\cdot 1_{\{y\neq y'\}}.
\end{equation}
Then, the key observation is that optimizing $\mathcal{L}_R$ can be solved by optimizing the Lagrangian relaxation with penalty parameter $\eta$:
\begin{equation}
    \mathcal{L}_{RL}:=\sup_P \{\E_P[l(\theta;(X,Y))] - \eta D_\theta (P, P_s)\}.
\end{equation}
The gradient of $\mathcal{L}_{RL}$, under a suitable condition, can be rewritten as \cite{volpi2018generalizing,boyd2004convex},
\begin{equation}
    \nabla_\theta L_{RL} =\E _{(X,Y)\sim P_s}[\nabla_\theta l(\theta; (x_\eta^*, Y))],
\end{equation}
where 
\begin{equation}
    x_\eta^* =\argmax_{x\in \mathcal{X}}\{l(\theta; (x,Y)) - \eta c_\theta ((x,Y)), (X,Y))\}.
\end{equation}
A min-max algorithm is used to estimate the gradients approximately as discussed in Sec.~\ref{sec:ada}.

\section{Additional Experimental Settings}

\begin{figure}[htp!]
\centering
\includegraphics[height=3.8cm]{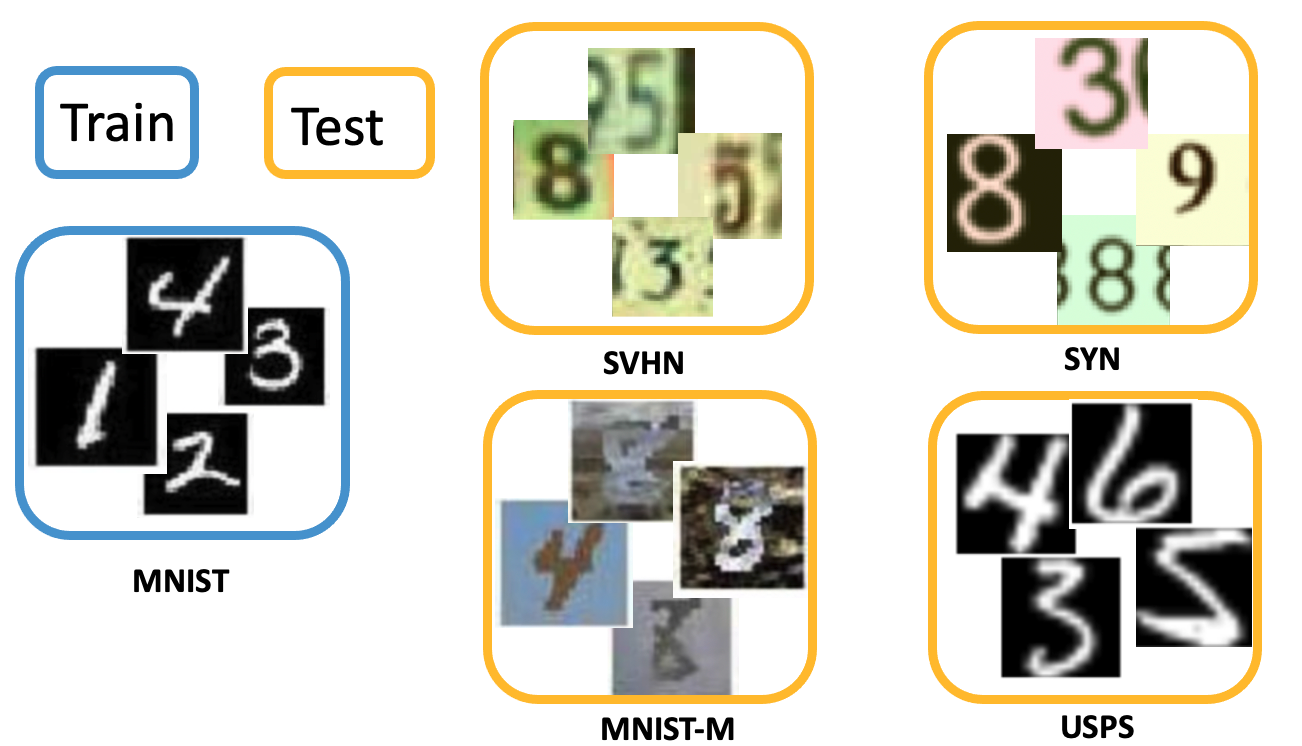}
\caption{\small Single domain generalization with Digits benchmark. Only MNIST is used for training and the goal is to learn a model that generalizes well to other digits domains, including, SVHN, MNIST-M, SYN, USPS. }
\label{fig:sdg} \vspace{-1.5mm}
\end{figure}

In Figure~\ref{fig:sdg}, we show some visual examples from the Digits benchmark. SVHN, MNIST-M and SYN are more challenging domains that have larger distributional shift from MNIST than USPS.

\end{document}